# Robust Detection of Non-overlapping Ellipses from Points with Applications to Circular Target Extraction in Images and Cylinder Detection in Point Clouds

Reza Maalek[1, *] and Derek D. Lichti[2]

[1] Endowed Professor of Digital Engineering and Construction, Institute for Technology and Management in Construction (TMB), Karlsruhe Institute of Technology (KIT). Address: Am Fasanengarten, Karlsruhe, Germany, 76131. Email: reza.maalek@kit.edu (corresponding author)

[2] Professor of Geomatics Engineering, Department of Geomatics Engineering of the Schulich School of Engineering, University of Calgary. Address: 2500 University Dr. NW, Calgary, Canada, T2N 1N4. Email: ddlichti@ucalgary.ca

* Indicates corresponding author

**Abstract:** This manuscript provides a collection of new methods for the automated detection of non-overlapping ellipses from edge points. The methods introduce new developments in: (i) robust Monte Carlo-based ellipse fitting to 2-dimensional (2D) points in the presence of outliers; (ii) detection of non-overlapping ellipse from 2D edge points; and (iii) extraction of cylinder from 3D point clouds. The proposed methods were thoroughly compared with established state-of-the-art methods, using simulated and real-world datasets, through the design of four sets of original experiments. It was found that the proposed robust ellipse detection was superior to four reliable robust methods, including the popular least median of squares, in both simulated and real-world datasets. The proposed process for detecting non-overlapping ellipses achieved F-measure of 99.3% on real images, compared to F-measures of 42.4%, 65.6%, and 59.2%, obtained using the methods of Fornaciari, Patraucean, and Panagiotakis, respectively. The proposed cylinder extraction method identified all detectable mechanical pipes in two real-world point clouds, obtained under laboratory, and industrial construction site conditions. The results of this investigation show promise for the application of the proposed methods for automatic extraction of circular targets from images and pipes from point clouds.

**Keywords:** non-overlapping ellipse detection, cylinder extraction, point cloud pipe detection, circular target extraction, smartphone camera calibration.

## 1. Introduction: automated ellipse detection

The reliable, automatic detection of non-overlapping ellipses is a common problem, especially for the calibration, registration, and geometric feature detection and modeling of optical instruments such as cameras and laser scanners. For systematic error modeling, the calibration of optical instruments often requires a well distributed planar circular target field with high redundancy (Fraser, 1997). Figure 1a, the calibration laboratory at the University of Calgary, illustrates an example of a permanent target field comprised of a total of 600 black and white planar circular targets of different sizes, which has been utilized extensively to calibrate various types of optical instruments such as terrestrial laser scanner (TLS) (Lichti et al., 2019), 3D range cameras (Lichti et al., 2010), and panoramic cameras (Lichti et al., 2020). One of the most important factors influencing the effectiveness of the calibration is the correct detection and accurate estimation of the center of these targets. Since the projective transformation of a circle onto a plane is approximated as an ellipse, the automatic, accurate and robust detection of non-overlapping ellipses from the edge points becomes a critical stage in the calibration process. Figure 1a-right shows an ideal ellipse detection output from image edge points, the results of which can be used for both calibration as well as registration of optical instruments.

Another prime example is the problem of cylinder detection in three-dimensional (3D) point clouds (an example is provided in Figure 1b). Since ellipse is the geometric construction (Gallian, 2021) of a cutting plane with a cylinder, intersection of a cutting plane with the cylindrical point clouds will produce points following elliptic patterns. Hence, the problem of cylinder detection from 3D point clouds can be reduced to detecting non-overlapping ellipses from two-dimensional (2D) points projected onto a cutting plane (Figure 1b-middle). The geometric parameters of the original cylinders, including the axis, center, and radius, can then be recovered from the detected ellipse parameters (Bergamasco et al., 2020; Miller and Goldman, 1992; Rahayem et al., 2012). The detected ellipses from the projected points of Figure 1b-middle are shown in Figure 1b-right, which can be then utilized to provide an initial estimate of the original cylinder's parameters. Once initial estimates of the cylinder's axis and radius are derived, the problem of multiple cylinder detection can be reduced to detecting one isolated cylinder from a set of neighboring points, which can then be solved using available and established methods such as the robust cylinder extraction of (Maalek et al., 2019), or the cylinder detection of (Tran et al., 2015).

This study presents a robust method for the detection of non-overlapping ellipses from 2D points, applicable to: (i) extracting circular targets from images, which is an important step in the calibration and registration of optical instruments; and (ii) detecting cylinders from point clouds, which plays an important role in the as-built building information modeling (BIM) of cylinder-like structures, such as mechanical pipes and columns. To detail the scope of this study more specifically, it is important to first introduce current methods of ellipse detection from 2D points along with some of their limitations in practical settings.

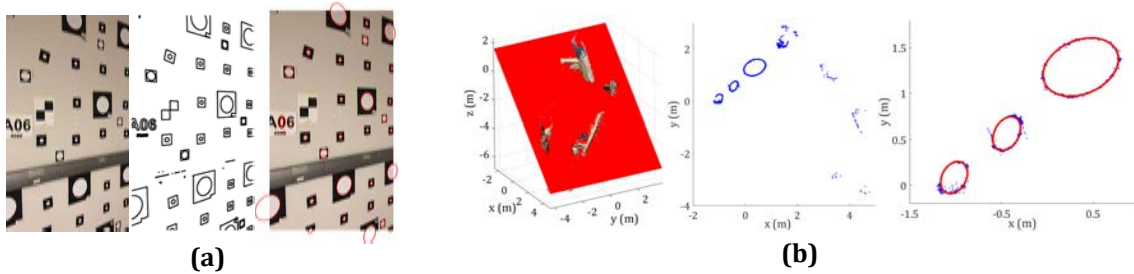

**Figure 1:** Use cases for detecting non-overlapping ellipses: a) detection of black and white targets for the purpose of instrument calibration and registration: image-left; image edge points-middle; detected ellipses-right; and b) detecting ellipses from projected 3D points on a plane (shown in red) for the purpose of cylinder extraction from point clouds: point cloud and intersecting plane-left; projected points onto the plane-middle; detected ellipses-right.

*1.1. Methods of ellipse detection from 2D points*

The problem of ellipse detection from 2D edge points[1] can be solved using one (or a combination) of the following methods:

1- Hough transform (HT), where an elliptical parameter sampling space (Hough space) is constructed and the parameters whose peak frequencies meet some predefined voting criteria are selected as ellipses.
2- Edge chaining or following, where potential elliptic arcs are first detected by analyzing the local behavior around each point. The detected arcs are then clustered together through some predefined grouping criteria. The clusters satisfying some elliptical validation criteria are selected as ellipses.
3- Top-down segmentation, where a connected component region growing method is first employed to determine connected regions. An ellipse is then fitted to each region and those regions conforming with some ellipse validation criteria are selected as elliptical.

HT-based methods require an efficient quantization procedure of the 5-dimensional parameter space (in the case of ellipses). In fact, the precision of the estimated ellipse parameters is a function of the size of the quantized parameter space cells (Illingworth and Kittler, 1988). With no available information of the ellipse parameters a priori, many parameter perturbations must, hence, be tested, which increase computational cost, especially in high resolution or complex scenes. To reduce computational cost, different variations of the original HT such as randomized HT (RHT; (Lu and Tan, 2008; McLaughlin, 1998; Xu et al., 1990)) and probabilistic HT (PHT; (Kiryati et al., 1991)) were introduced, which analyze only a subset of the edge candidates. However, as noted by (Mai et al., 2008), the RHT can lose efficiency as the number of ellipses increase, and the PHT assumes a uniform distribution of the edge points, which is impractical for real data. On the other hand, in rare cases where a priori knowledge of the ellipses exists, one can reduce the search domain and the dimensions of the parameter space (e.g. similar to the cylinder detection in (Rabbani and Van Den Heuvel, 2005). For instance, (Yonghong and Qiang, 2002) proposed a simple method that only requires a one-dimensional accumulator array (as opposed to five) for the semi minor axis, given two points on the major axis. Since two points on the major axis of an ellipse are required (through pair-wise testing), the method, however, is limited to only arc lengths of larger than a semi-ellipse.

---
[1] Here, pixel edges of images are also treated as points.

Edge chaining methods, on the other hand, are more suited for detecting a larger number of overlapping ellipses with smaller arc lengths. The methods start by examining the neighborhood of each edge pixel to determine its local behavior. A region growing method is then employed to group together the adjacent points satisfying some geometric constraint, such as gradient smoothness, connectivity, and convexity. The segmented arcs that potentially form the same ellipse are then clustered together and the final set of ellipses are determined based on some validation criteria. These methods, similar to any region growing method (Maalek et al., 2018; Rabbani et al., 2006), however, require efficient tuning of different thresholds of parameters such as, neighborhood size, gradient angle change, and ellipse validation criteria. In addition, the clustering of arcs requires an accurate method to estimate the center of an ellipse. In fact, in small arc lengths (say quarter ellipse), the center estimation becomes biased using even the most reliable ellipse fitting methods (Maalek and Lichti, 2020). Researchers have since proposed alternative methods to provide an estimate of the center of an ellipse (Ser and Siu, 1995; Yin, 1994). However, these center estimations are impacted by noise (Prasad et al., 2012), requiring another threshold to account for the center estimation uncertainty between potential arcs of the same ellipse. In line with threshold tuning, the methods proposed by (Meng et al., 2020), (Fornaciari et al., 2014), (Lu et al., 2020), (Dong et al., 2018), and (Liu et al., 2019) require the tuning of thresholds for four, five, seven, eight, and twelve parameters, respectively, all of which must be determined empirically[2]. In fact, the suitability of a given threshold for all types of images cannot be guaranteed (Dong et al., 2018), which impacts the generic applicability of the method. Nevertheless, the edge chaining methods remain the state-of-the-art when detecting complex, overlapping, and occluded elliptical arcs from edge points.

Another class of ellipse detection method, referred to here as top-down segmentation, aims to detect large numbers of non-overlapping ellipses from edge points, such as the calibration problem shown in Figure 1a. A connected component region growing is first performed on the edge points to determine the connected regions. Since the ellipses are assumed non-overlapping, each ellipse should theoretically occupy only one connected region (in practice, this may not necessarily hold as can be seen in Figures 2a and 2b). An ellipse is then fitted to the connected regions. Based on some validation criteria, the elliptical regions are then detected. For instance, (Jarron et al., 2019) used predefined threshold ranges for the expected aspect ratio and size of the ellipses as validation criteria. This thresholding, however, can only detect ellipses of limited sizes and aspect ratios, which often change between different scenes. (Pătrăucean et al., 2017) used an *a contrario*-based validation strategy to differentiate between linear and elliptical regions in a dataset. Another more exotic ellipse detection method, proposed by (Panagiotakis and Argyros, 2020), used Akaike information criterion (AIC) to optimally detect the ellipses in a given connected shape (suitable for overlapping and non-overlapping ellipses). In our datasets, however, the latter two methods failed to classify many elliptic regions (Type I error) and detected some non-elliptic regions as elliptical (Type II error; see Figures 18 and 19). To this end, a successful and generic implementation of the top-down segmentation process requires a stable and reliable validation metric and criteria.

---

[2] The reader is referred to the cited manuscripts for discussions on their empirical threshold tuning.

Another important data artefact is the possible existence of outliers in connected edge points. Figure 2 illustrates examples of outliers in connected points of both images and projected point clouds. In images, blurring and lighting conditions can negatively impact the produced binary image and consequentially the detected edges (Figures 2a and 2b). Figure 2a demonstrates an example where the ellipse and the rectangular background (of one of the targets) are merged into one segment after connected components. Figure 2b shows the further division of a connected ellipse into two separate regions (only one region is shown in this example). Figure 2c shows a set of connected projected points from a cylinder and its vicinity. In this example, the point cloud was acquired from mechanical pipework at a construction site, where data artifacts due to occlusion, dust and movement had produced outliers (as can be observed in the image). In these cases, even if effective and robust ellipse validation criteria exist, points following elliptic patterns must first be detected from each connected segment since the combined segment clearly does not follow an elliptic pattern. As a point of reference, the result of robust ellipse fitting (red ellipse) using our new method -to be described in Section 3.5- is compared to the algebraic ellipse fit of (Kanatani and Rangarajan, 2011) (green ellipse). For instance, the center of the best fit ellipse estimated in Figure 2b (green) is clearly and significantly different from the true center of the ellipse (about 50 pixels difference for an ellipse with semi-major length of around 50 pixels). Furthermore, in Figure 2c, the angular error between the estimated and original cylinder's axis increased from 4° to roughly 30° using our robust method and Kanatani's ellipse fit, respectively. Therefore, ellipse fitting to connected segments with outliers will clearly not produce favorable results, and a robust strategy must be employed to first detect elliptic points from the connected segments.

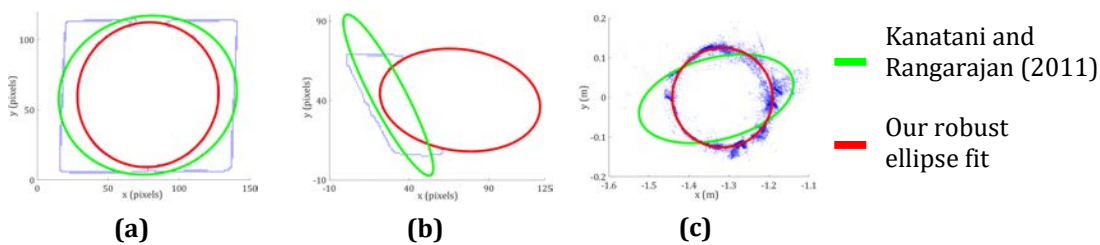

**Figure 2:** Example of connected segments with outliers- top; and results of robust ellipse fitting (red) and regular ellipse fitting (green)-bottom: a) image edge points of a complete ellipse with outliers; b) image edge points of an elliptic arc with outliers; c) point cloud of a cylinder projected onto a plane

*1.2. Objectives and scope of this study*

Considering the main limitations, outlined at the end of Section 1.1, the main objective of this study is to propose a generic method for the robust fitting and detection of non-overlapping ellipses from 2D edge points by proposing: (i) a generic, systematic and effective ellipse validation criterion to differentiate between connected points following elliptic and non-elliptic patterns; and (ii) a robust ellipse detection and fitting method to extract elliptic points in the presence of outliers. The effectiveness of the proposed methods will be validated in practical applications pertaining to circular target extraction from images and cylindrical mechanical pipe detection from point clouds.

*1.3. Structure of the manuscript*

The remainder of the manuscript is structured as follows. Section 2 presents the review of the state-of-the-art robust curve fitting approaches as well as substantiated elliptical shape descriptors. In Section 3, the detailed explanation of the proposed algorithms for robust ellipse fitting, and ellipse detection, followed by the methods and metrics to validate the proposed algorithms are provided. Section 4 outlines the design of the simulated and real-world experiments, carried out to assess the applicability of the proposed methods in both images and point clouds. In Section 5, the results of the designed experiments are reported. Section 6 provides the summary of the methods along with major findings of this study. Section 7 presents some avenues for further investigations.

## 2. Background

### 2.1. Robust ellipse fitting

For the problem of robust curve fitting, the most widely deployed and practical algorithms are based on the Monte Carlo method (Zhang, 1997). The Monte Carlo curve fitting idea is predicated on generating and testing many subsamples of k points from all data points ($n \gg k$) with hopes that at least one outlier-free subsample is chosen. The subsample containing no outliers (outlier-free) can then be used to determine other inliers in the data and estimate the best fit curve parameters. The more samples drawn, the higher the chance of obtaining at least one outlier-free subsample will be. Hence, the following equation is commonly used to determine the minimum number of points required to draw at least one outlier-free subset of $k$ points, given an outlier ratio, $\epsilon$, and probability, $P$:

$$N = \frac{\log(1-P)}{\log(1-(1-\epsilon)^k)} \tag{1}$$

The main difference between most Monte Carlo-based approaches is the choice of sample size, $k$, and the validation criteria used to detect inliers. Here, we present three of these methods, namely, random sample and consensus (RANSAC (Fischler and Bolles, 1981)), least median of squares (LMedS (Rousseeuw and Leroy, 1987)) and least trimmed squares (LTS (Rousseeuw and Van Driessen, 2006)), which have been widely implemented for robust ellipse (Rosin, 1999; Zhang, 1997) and circle (Maalek et al., 2019; Nurunnabi et al., 2018) fitting.

RANSAC methods typically use the minimum sample size required to provide an estimation of the model parameters, which depending on the ellipse fitting method is either 5 or 6. For each subsample, the distance of all points to the fitted ellipse is then estimated using one of the available point to ellipse distance functions (e.g. algebraic (Paton, 1970), Sampson (Sampson, 1982), confocal hyperbola (Maalek and Lichti, 2020; Rosin, 1998), geometric (Chernov and Wijewickrema, 2013), and etc.)[3]. Assuming data points with Normally distributed random errors, $\mathcal{N}(0, \sigma^2)$, the inliers are then determined using the following equation:

$$\begin{cases} \text{inlier}, & d_i^2 < \chi_{q,\alpha}^2 \sigma^2 \\ \text{outlier}, & d_i^2 \geq \chi_{q,\alpha}^2 \sigma^2 \end{cases} \tag{2}$$

where $d_i$ is the distance of point $i$ to the fitted ellipse, $\chi_{q,\alpha}^2$ is the chi-squared cumulative probability distribution function with probability $\alpha$ and degrees of freedom $q$ (2 for 2D data), and $\sigma$ is the standard deviation of the observations in $x$ and $y$ (assuming independent and identically distributed). The subsample containing the largest number of inliers (most consensus) is then selected as the final set. In the case of equal number of inliers, the subset with the lowest sum of squared distances is typically selected (Hartley and Zisserman, 2004). The main shortcoming of RANSAC is that the variance of the observations is not generally known a priori. If the variance is selected too large, the highest consensus set will contain outliers (Type II errors). Alternatively, if the variance is selected too small, many inliers will be detected as outliers (Type I errors) or the algorithm runs the chance of detecting no inliers.

---

[3] Based on the results of (Maalek and Lichti, 2020), our choice is the confocal hyperbola distance since it accurately predicts the ground truth geometric distance, while producing significantly faster distance approximations.

LMedS, on the other hand, introduces an alternative approach to the adaptive estimation of the variance. First, amongst all subsamples, the one with the smallest median of its squared distances is selected. The standard deviation, $\hat{\sigma}$, can then be estimated using one of the following robust methods:

$$\begin{cases} \text{Rosin [36]} & \hat{\sigma} = 1.4826\left(1+\frac{5}{n-1}\right)C_R \text{ with } C_R = \underset{i=1,\dots,n}{\text{median}}\left|d_i - \underset{i=1,\dots,n}{\text{median}}(d_i)\right| \\ \text{Zhang [32]} & \hat{\sigma} = 1.4826\left(1+\frac{5}{n-k}\right)C_Z \text{ with } C_Z = \sqrt{\underset{i=1,\dots,n}{\text{median}}\, d_i^2} \end{cases} \quad (3)$$

where |.| denotes the absolute value function. The main drawback of LMedS is that the ratio of outliers cannot exceed 50% ($\frac{\left\lfloor\frac{n-k}{2}\right\rfloor+1}{n}$ to be exact); however, it is possible to altogether replace the median with an appropriate percentile (Zhang, 1998) -say 25th percentile for 75% or less outlier ratio. The solution, however, may not be globally optimal and the efficiency of the standard deviation estimates of equation 3 cannot be guaranteed.

Another method that can achieve higher efficiency compared to LMedS, especially in larger datasets is the LTS algorithm (Davies, 1990). Currently, only three algorithms exist that can provide an exact solution to LTS, and the most computationally efficient one (Klouda, 2015) requires around 500,000 initial subsamples to find the LTS for a circle fitting problem with only 50 data points (Maalek et al., 2019). As a result, the approximate solution provided in (Rousseeuw and Van Driessen, 2006), called Fast-LTS, is currently the most practical algorithm available. Fast-LTS starts by selecting many subsamples of size, $k = n(1-\epsilon)$. For each subsample, a concentration step (Rousseeuw and Van Driessen, 1999) is performed to determine the $k$ samples that best comply with the model (i.e. the inliers). The final set of inliers belong to the subset whose sum of the squared distances of the $k$ identified points is the least. Fast-LTS has shown to be effective in circle fitting (Nurunnabi et al., 2018); however, it requires a correct initial outlier ratio to estimate the sample size $k$. Furthermore, similar to LMedS, it can only handle up to 50% outliers. To improve on some of the shortcomings associated with Fast-LTS, (Maalek et al., 2019) proposed some fundamental adjustments to the concentration step, which significantly improved the results and dependency of the algorithm on the outlier ratio. One notable improvement was the use of mode instead of robust mean for the intercept adjustment, which allowed for more flexibility and accuracy, especially in outlier ratios above 50%. These adjustments improved the overall accuracy of the inlier detection for circle extraction from 86.92% to 99.23%. To this end, this latter method will be deployed in this study to aid with robust ellipse fitting (see Section 3.5.2 for more details).

Another important consideration when applying Monte Carlo-based methods for robust ellipse fitting is the geometry of the $k$ samples. This is since the arc length of an ellipse covered by the observations plays a crucial role in the accuracy of the estimated ellipse parameters. To this end, every effort must be made so that the initial $k$ samples are not adjacent and cover the whole range of the ellipse, otherwise, the samples will be practically useless. (Rosin, 1999) showed that random sampling will result in a high probability of obtaining point sets with directly adjacent points. In fact, for a set of 20, 30 and 100 points, there is about 80%, 50% and 20% chance, respectively, that a selection of five points contains exactly adjacent points (see Fig.4a of (Rosin, 1999)). To prevent the selection of adjacent points, (Zhang, 1997) proposed a bucketing strategy, whereby the data are first divided into rectangular grids, called buckets. The problem is now reduced to selecting $k$ buckets and then further selecting one point from each bucket. The strategy showed great promise for the estimation of fundamental matrices from sets of matching points in two images (Zhang, 1998). However, the method still relies on a subjectively defined bucket grid size. Furthermore, the rectangular grids are not geometrically suited to ensure the entire elliptic arc (or a good portion of it) is covered. To this end, a new bucketing strategy, specifically for robust ellipse fitting, will be proposed that guarantees to return sample points over the elliptical arc, while eliminating the chance of selecting two adjacent points in one subset (see Section 3.5.1).

*2.2. Measuring ellipticity*

After an ellipse is fitted to a set of connected points, it becomes important to determine whether the points follow an elliptic pattern or not. For this, various metrics, called ellipticity measures, provide an indication of whether a set of points closely follow an elliptical pattern. One of the most complete studies in the domain of ellipticity measures was presented by (Rosin, 2003), where various metrics were discussed and compared. Amongst the five ellipticity metrics introduced in (Rosin, 2003), three that performed well, namely, elliptic variance ($E_V$ (Peura and Iivarinen, 1997)), moment invariants ($E_I$)[4], and Euclidian ellipticity ($E_E$), will be considered in this study. The results of all three methods are normalized to range from least elliptical, zero, to most elliptical, one.

Elliptic variance ($E_V$) is calculated as follows:

$$E_V = \frac{1}{1+\sigma_V} \text{ with } \sigma_V = \frac{\sum_{i=1}^{n}(r_i-M_r)^2}{nM_r^2} \text{ and } M_r = \frac{\sum_{i=1}^{n}r_i}{n} \tag{4}$$

where $r_i$ is the Mahalanobis distance of point $i$ from the mean, $M_r$ is the mean of the Mahalanobis distances, and $n$ represents the number of points. In general, the locus of points in a plane with the same Mahalanobis distance from the mean is represented by an ellipse. In the case of uniformly distributed points on a complete ellipse, the Mahalanobis distance is expected to provide the equation of the ellipse. For an incomplete ellipse, say a half ellipse, however, since the mean of the data shifts, the Mahalanobis distance from the mean no longer represents the equation of the best fit ellipse. Therefore, the elliptic variance is expected to only perform well for points of complete ellipses with close to uniform distribution.

Moment invariants ($E_I$) is calculated as follows:

$$E_I = \begin{cases} 4\pi^2\sigma_I^2, & \sigma_I^2 \leq \frac{1}{4\pi^2} \\ \frac{1}{4\pi^2\sigma_I^2}, & \sigma_I^2 > \frac{1}{4\pi^2} \end{cases} \text{ with } \sigma_I^2 = \mathcal{H}_1^2 - \mathcal{H}_2 \tag{5}$$

where $\mathcal{H}_1$ and $\mathcal{H}_2$ are the first and second Hu moments (Hu, 1962) of the shape enclosed by the edge points. An important consideration for practical implementation of moment invariants is that it cannot be applied to only edge points. A hollow elliptical shape has an $E_I$ close to zero (see Fig. 3 of (Žunić et al., 2017)); hence, in images, all pixels inside of a given boundary point must be considered in the computation of $E_I$, which of course increases the computation time. This approach also becomes less practical for projected point clouds since the projected points are not gridded naturally (unlike digital images), and the choice of points enclosed by the boundary points becomes arbitrary. Similar to $E_V$, $E_I$, by definition, is only suited for complete elliptical shapes ((Rosin, 2003) pp. 173), and half ellipses produce a smaller $E_I$. To this end, the value of the invariant (i.e. $4\pi^2$) must be adjusted accordingly to accommodate smaller elliptical arcs.

Euclidean ellipticity ($E_E$) is calculated as follows:

$$E_E = \frac{1}{1+\frac{\sigma_E}{\sqrt{A}}} \tag{6}$$

---

[4] $\sqrt{E_I}$ is referred to as the maximum ellipticity (Žunić et al., 2017), which had shown success in galaxy classification from images.

where $\sigma_E$ is the root mean squared error of the boundary points to the best fit ellipse, and $A$ is the area of the best fit elliptical arc. Amongst the three ellipticity measures $E_E$ is the only one that incorporates the deviations of the points from the best fit ellipse within the ellipticity measure; furthermore, $E_E$ can be computed directly from the boundary points without the need to construct shape. In Section 5.2.1, these three metrics will be compared to find the most suitable measure capable of differentiating between ellipses and rectangles.

## 3. Methodology

As stated in Section 1, this manuscript focuses on the robust detection of non-overlapping ellipses from edge points with application to circular target extraction from images and cylinder detection from point clouds. The proposed solution is formulated as follows:

1- **Pre-processing:** converting the image and point cloud into 2D edge points:
   a) **Image pre-processing:** converting the color image (Figure 3a) into image edge points (Figure 3b).
   b) **Point cloud pre-processing:** generating 2D points projected onto a plane (Figure 4a-c).
2- **Connected component labeling:** segmenting the connected edges (Figures 3c and 4d).
3- **Ellipse fitting:** ellipse fitting to each connected segment (Figures 3d and 4e).
4- **Ellipse validation:** detection of connected edges following elliptical pattern (Figure 3e).
5- **Robust ellipse fitting and validation:** robust ellipse fitting on the following remaining regions (in order):
   a) unvalidated connected regions whose best fit ellipse does not enclose the center of an existing validated ellipse (Figures 3f).
   b) combined validated ellipses of steps 4 and 5a whose best fit ellipses overlap (Figure 3g).
6- **Post-processing for point clouds:** detecting cylinders from extracted ellipses (Figure 4f).

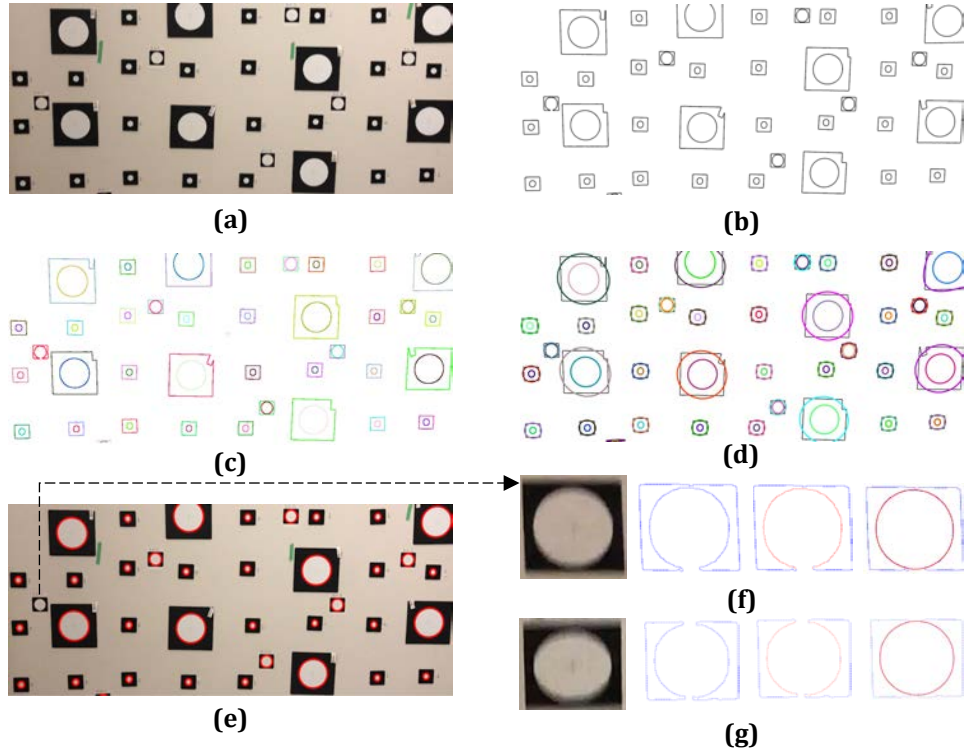

**Figure 3:** Overview of the proposed method for automatic detection of non-overlapping ellipses from images: a) Original image; b) detected edges; c) connected component segmentation (each color represents one segment); d) best fit ellipses to each connected segment; e) detected ellipses in red satisfying the validation criteria; robust ellipse fitting on segments with: f) rectangles and ellipses incorrectly grouped; and g) rectangles and ellipses incorrectly divided

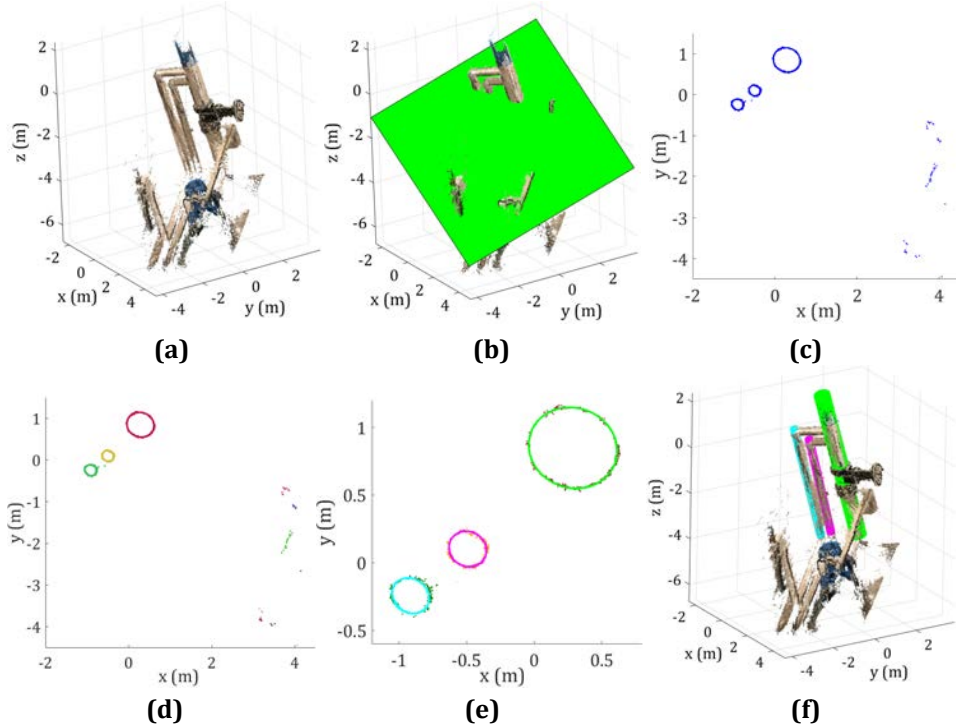

**Figure 4:** Overview of the proposed method for automatic detection of cylinders from point clouds: a) Original point cloud; b) point cloud intersected by a plane (shown in green); c) projected points onto the plane; d) connected component segmentation (each color represents one segment); e) detected ellipses satisfying the validation criteria; and f) set of extracted cylinders corresponding to the detected ellipses

## 3.1. Pre-processing

This stage is used to convert the image and point cloud into 2D edge points. Since the nature of generating 2D edge points from images and point clouds are different, they will be explained separately in the following.

### 3.1.1. Image pre-processing

For images, the pre-processing stage consists of three main steps, namely, conversion of RGB to gray, conversion of gray to binary, and edge detection from binary. Various methods exist to convert image RGB values to grayscale (Kanan and Cottrell, 2012; MacEdo et al., 2015). In this study, the luminance method was used, which approximates the image's gradient to account for humans' brightness perception. Binarization is then carried to convert the greyscale intensities into zeros or ones by using either a fixed global threshold (Otsu, 1979), or an adaptive local threshold. Since adaptive thresholds can better account for local variations in an image, here, the adaptive thresholding method of (Bradley and Roth, 2007) is employed, which was shown to perform well in binarizing black and white augmented reality targets. Finally, the edges are detected using the method of (Canny, 1986), which had shown promising results in detecting edges of black and white circular targets from laser scanner intensity images (Lichti et al., 2019).

### 3.1.2. Point cloud pre-processing

To project 3D points onto a plane, a simple parallel projection is carried out, whereby points within a pre-defined distance from the plane are projected in the direction of the plane's normal. This can be intuitively accomplished by rotating the data such that the plane's normal is parallel to the $z$-axis (i.e. $(0,0,1)^T$), and then selecting the $x$-$y$ coordinates of points within the pre-defined distance from the plane. It is worth noting that the location and orientation of the intersecting plane, the resolution of the point cloud, along with the pre-defined distance of points to the plane play key roles in the ellipse detection and consequentially the cylinder extraction processes. For instance, with no a priori knowledge of the number and orientation of cylinders, finding the minimum number of planes required to pass through all cylinders is not a trivial problem. Even when a plane is so chosen to pass through a cylinder, missing data due to occlusions, or low resolution of the point cloud at the vicinity of the plane may prevent the formation of elliptic patterns, in which case an ellipse and consequentially a cylinder will not be detected. Our goal here, however, is to examine the possibility of detecting cylinders from point clouds, when a plane passing through the cylinders is known a priori, and the projected points form elliptic patterns. In this manuscript, the planes were selected manually and the points within 1mm of the plane were chosen as the final 2D points.

*3.2. Connected component labeling*

In this stage, connected image edge pixels and projected 2D point clouds are grouped together using a simple region growing method. For images, the closest eight pixels (8-connectivity) to a given point, corresponding to points within $\sqrt{2}$ pixel radius, is used. For the projected point clouds, however, defining the connected component radius is not as simple. This is attributed to the fact that the projected points are not naturally in gridded form and hence, the distance between connected points varies. To accommodate for the variability of the distance between points, the adaptive neighbourhood definition of (Weinmann et al., 2015) is adopted. The method was proposed for 3D, but can also be used in 2D, and it requires a minimum and maximum number of points or radii range as input. Here, the radius of points is changed from 1cm as the minimum to 15cm as the maximum in 1cm increments. Once the optimal radius for each projected point is defined, a region growing method is employed starting from any seed point to group together all connected points (Figures 3d and 4d).

*3.3. Ellipse fitting*

Once the connected points are identified, an ellipse is fitted to each segment. There are numerous ellipse fitting methods that minimize different distance functions, such as algebraic (Fitzgibbon et al., 1999; Halir and Flusser, 1998), Sampson (Kanatani and Rangarajan, 2011; Szpak et al., 2015), and geometric (Ahn et al., 2001). In (Maalek and Lichti, 2020), these methods were compared to the newly proposed confocal hyperbola-based ellipse fitting method. It was demonstrated in both simulated and real-world data, that the confocal hyperbola performed close to identically to (and in many cases better than) the iterative geometric method of (Ahn et al., 2001), and outperformed the remaining methods, while achieving computation times closer to direct methods. Therefore, the confocal hyperbola-based ellipse fitting was used throughout this study to fit ellipses to data.

*3.4. Ellipse validation criterion*

As stated in Section 2.2, once an ellipse is fitted to the connected regions, a validation criterion is required to distinguish between ellipses and other shapes such as rectangles. Here, the Euclidian ellipticity measure introduced in equation 6 is utilized to detect ellipses as follows:

$$\begin{cases} \text{ellipse} \quad , & E_E > Th_E \\ \text{not ellipse} \, , & E_E \leq Th_E \end{cases} \quad (7)$$

where $Th_E$ is the threshold used to distinguish between elliptical and non-elliptical segments. The Euclidian ellipticity measure is attractive since it utilizes the root mean squared error of the best fit ellipse normalized by the square root of the ellipse's area and can be applied directly on the boundary points (no need to generate shape). In Section 5.2, the three ellipticity measures introduced in Section 2.2 were compared, and the value of $Th_E = 0.96$ was derived. In our simulations and real-world experiments, the Euclidian ellipticity using the empirically derived threshold of 0.96 consistently differentiated ellipses from rectangles.

*3.5. Robust ellipse fitting*

As shown in Figure 2, when a connected segment contains outliers, the fitted ellipse and consequentially the detection of elliptic segments will be impacted. Here, a new Monte-Carlo based method for robust extraction of points following elliptical patterns in the presence of outliers is introduced. Following the discussions presented in Section 2.1, the proposed Monte Carlo-based method is designed to address the following three main problems:

1- **Method of sample selection** to ensure samples do not contain adjacent points. Here, a new bucketing strategy based on the polar angle of points is developed.
2- **Inlier detection method** to determine the inlier points for each subsample. Here, an affine transformation is first performed to reduce the ellipse detection into a circle detection problem. The robust circle detection of (Maalek et al., 2019) is then employed to determine the inlier points for each subsample.
3- **Validation criteria** for selection of the best inlier set amongst all subsamples. Here, the thresholds and criteria are adaptive, systematically derived, and generalizable to different scenes and pixel noise levels.

The following subsections will detail the methods developed to address these main concerns.

*3.5.1. Method of sample selection*

As explained in Section 2.1, the geometry of the samples plays a crucial role in the correct estimation of ellipse parameters. If all points are adjacent and only cover a small fraction of the ellipse, the results of the ellipse fitting will surely be inaccurate and consequentially the subsample will be useless. The proposed bucketing strategy must, hence, reduce (or in our case eliminate) the possibility of selecting adjacent points within the subsamples while providing a good distribution of points along the elliptic arc. To this end, we propose a bucketing strategy based on polar angles of the points with respect to the center of the best fit ellipse. Once the polar angles (from 0 to $2\pi$) of the points are determined, the polar angle space is divided evenly into $2p$ sections ($p \geq k$ = number of samples). The problem is now reduced to randomly selecting $k$ sections from two alternating $p$ sections. We then randomly select one point from each selected section. This process, in fact, guarantees that no two adjacent sections and consequently no two adjacent points are selected in one subsample. To formulate the process, **Algorithm 1: Method of Sample Selection** is developed as follows:

1- Estimate the geometric parameters of the best fit ellipse $\rho = (x_c, y_c, a_e, b_e, \theta)^T$, where $(x_c, y_c), a_e, b_e$ and $\theta$ are the center, semi-major length, semi-minor length, and rotation angle, respectively (Figure 5a).
2- Apply the following affine transformation to the data points to convert the best fit ellipse to a circle with radius, unity, and center, (0,0):

$$\begin{bmatrix} x \\ y \end{bmatrix} = \begin{bmatrix} \frac{1}{a_e} & 0 \\ 0 & \frac{1}{b_e} \end{bmatrix} \begin{bmatrix} \cos(\theta) & \sin(\theta) \\ -\sin(\theta) & \cos(\theta) \end{bmatrix} \begin{bmatrix} X - x_c \\ Y - y_c \end{bmatrix} \tag{8}$$

where $(X, Y)$ and $(x, y)$ are the original and the transformed point coordinates, respectively (Figure 5b).

3- Estimate the $2\pi$ angle (full circle angle) of the points in the new $(x, y)$ coordinate system. Sort the angles and find the largest difference (gap) between two consecutive angles in the counterclockwise direction (say, $\alpha_2 - \alpha_1 = \alpha$). Define the arc angle as $\beta = 2\pi - \alpha$ and shift all polar angles by $\alpha_2$ so that the shifted angle of $\alpha_2$ becomes zero and $\alpha_1$ becomes $\beta$ (Figure 5c).

4. Divide the new shifted angles into $2p$ sections ($p \geq k$) of arc angle $\frac{\beta}{2p}$ (starting from angle zero; Figure 4d). Separate the sections into two subsets of $p$ sections, one containing ordered odd sections, $1,3 \ldots 2p-1$, and the other even sections, $2,4 \ldots 2p$ (Figure 5e).
5. Estimate the number of subsamples, $N$, using equation 1. To select the same number of subsamples for both subsets of $p$ sections (odds and evens), $N = 2\left\lceil \frac{N_{eq.\,1}}{2} \right\rceil$ is chosen, where $\lceil . \rceil$ denotes the ceiling function.
6. Independently select $\frac{N}{2}$ subsamples of $k$ random sections from the odd and even sections, respectively.
7. From each of the selected $k$ sections of the $N$ subsamples, randomly choose one point.

The last variable that must be defined is the divisor, $p$. If the selected $p$ is too large, the $k$ samples may still cover only a small arc section, which is again unfavourable. Here, we define $p$ with respect to a user specified arc angle, $\varphi \leq \beta$, that must be covered if $k$ consecutive odd (or even) sections are selected:

$$p = \left\lceil \frac{k\beta}{\varphi} \right\rceil \tag{9}$$

Steps 1-4 of Algorithm 1 are schematically shown in Figure 5. The algorithm ensures that no two adjacent sections and consequentially points are selected in one subsample.

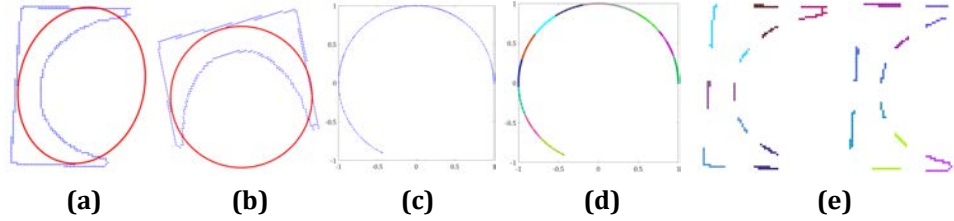

(a) (b) (c) (d) (e)

**Figure 5:** Schematic representation of the sampling strategy of Algorithm 2: a) best fit ellipse to the connected points (in red); b) affine transformation of the data (equation 8); c) shifted full-circle polar angles of the points; d) arc angles divided into $2p$ evenly spaced angles, each color represents a segment of points (here, $p$ is set as 6 only for demonstration purposes); e) the two separate sets of $p$ sections

*3.5.2. Inlier detection method*

Now that a process to effectively select $k$ samples is in place, the remaining data points complying with the elliptical pattern of these samples must be determined. To this end, first, the best fit ellipse to the set of $k$ sample points is estimated and then the transformation of equation 8 is applied to convert the ellipse into a circle with radius, unity, and center at the origin. The problem is now reduced to finding a circle from the set of points. Converting the ellipse detection to a circle detection in this manner has two main merits. First is that accurate and established non-iterative circle fitting methods (Al-Sharadqah and Chernov, 2009; Pratt, 1987) already exists, which during the concentration steps will yield a reliable fit with better computational efficiency compared to the iterative methods (Maalek et al., 2019). Secondly, the transformation suggests that if the initial $k$ samples are all inliers (i.e. from the desired ellipse), the final set of extracted circular points (after transformation) will also form a circle with radius close to unity and center close to the origin. This latter property allows for the definition of an efficient, parameter-less, and normalized selection criteria between the different subsamples (see Section 3.5.3). To this end, the inliers are determined using **Algorithm 2: Inlier Detection Method** as follows:

1. Using the $k$ samples (red points in Figure 6a), estimate the geometric parameters of the best fit ellipse (Figure 6b).
2. Apply the affine transformation of equation 8 to all data points (Figure 6c) to convert the best fit ellipse to a circle with radius, unity, and center, (0,0).

3- Perform the following concentration step using Algorithm 1 of (Maalek et al., 2019) for robust circle fitting on the transformed points as follows (output of this step shown in Figure 6d):
   a) Calculate the squared distance, $d_i^2$ of each point, $i$, to the center of the circle; keep in mind that from step 2, the initial center is the origin (0,0).
   b) Estimate the mode of $d_i^2$, $Mod$ to determine the intercept adjustment.
   c) Estimate the standard deviation using the normalized median absolute deviation ($MADN$; (Hampel, 1974)) of $d_i^2$:

$$MADN = \frac{\underset{i=1,\dots,n}{\text{median}}\left|d_i^2 - \underset{i=1,\dots,n}{\text{median}}(d_i^2)\right|}{0.67449} \tag{10}$$

   d) Find all points satisfying the following condition:

$$\frac{|d_i^2 - Mod|}{MADN} \leq \sqrt{\chi^2_{0.95,2}} \approx 2.45 \tag{11}$$

   e) Consider the inliers as the new set of points:
   - If the inlier points remain unchanged between two consecutive iterations–or the objective function of the circle fit yields zero, which is rare in practice with measurement error–exit the concentration step and return the final set as the inliers. Calculate the radius and center of the final set of inliers using the hyper-accurate fit (Al-Sharadqah and Chernov, 2009) (the latter information will be used in Algorithm 3).
   - Else, calculate the center of the new set of points using the (Pratt, 1987) circle fit and return to step a.

4- Fit an ellipse to the final set of inliers from step-3 (in the original coordinate system) and estimate the point distances to the ellipse as the residuals[5], $Res_i$.

5- Perform Algorithm 2 of (Maalek et al., 2019) to determine all points complying with the inlier set as follows (output of this step shown in Figure 6e):
   a) Calculate the sample standard deviation of the residuals, $\sigma_f$, using the final set of inlier points.
   b) Find all points satisfying the following equation:

$$\frac{|Res_i|}{\sigma_f} \leq \sqrt{\chi^2_{0.95,2}} \tag{12}$$

The final set of points identified by step 5-b, $in_e$, is the set of detected inlier points following an elliptical pattern. Figure 6 provides a schematic representation of the steps carried out in Algorithm 2.

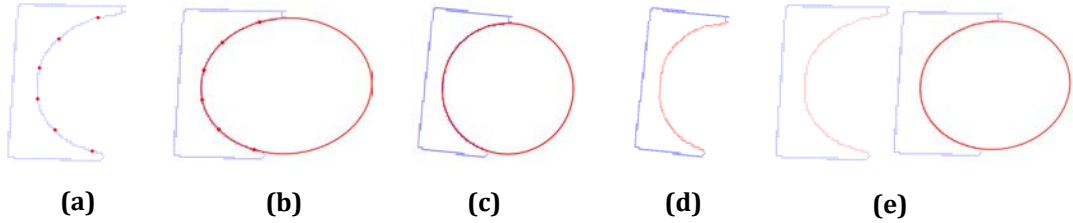

(a)      (b)      (c)      (d)      (e)

**Figure 6:** Schematic representation of the proposed robust ellipse fitting method: a) $k$ samples shown in red points; b) best fit ellipse shown in red solid line; c) affine transformation of the ellipse using equation 8; d) extracted points following circular pattern after concentration step; e) robust detection of the final elliptical inliers in the original coordinate system (extracted points and ellipse in red).

---

[5] The confocal hyperbola distance approximation is used since it is fast and accurately predicts the geometric distance (Maalek and Lichti, 2020).

*3.5.3. Validation criteria*

After the inlier set for each subsample is determined, the best subsample amongst all subsamples must be identified. To this end, the properties of the affine transformation of step 2 of Algorithm 3 are used. In fact, if the $\{k\}$ samples were inliers, the final set of inlier points in the transformed coordinate system would also: (i) contain the initial $\{k\}$ samples; (ii) have a radius, $r_s$, close to unity; and (iii) have a center, $(x_s, y_s)$, close to the origin. These criteria combined with the ellipse specific validation criterion presented in equation 7 are used to determine the best subset using **Algorithm 3: Robust Validation Criteria**, as blow:

1- Find all subsamples, satisfying the condition: $\{k\} \backslash in_e = \emptyset$, where $A \backslash B$ represents the set difference between sets $A$ and $B$ (also known as only $A$), $in_e \neq \emptyset$ is the final set of inlier points from Algorithm 2, and $\emptyset$ represents the empty set. If no set satisfies this condition, no ellipse is determined, and exit the algorithm.
2- Find all remaining subsamples, satisfying the threshold of equation 7. If no set is obtained, no ellipse is determined, and exit the algorithm.
3- From the remaining $M$ subsamples, select the subsample whose best fit circle parameters $(r_s, x_s, y_s)_i$, obtained in step 3-e of Algorithm 2 is closest to $(1,0,0)$, i.e.:

$$\underset{i=1,\ldots,M}{\mathrm{argmin}} \left\| \begin{bmatrix} r_s - 1 \\ x_s \\ y_s \end{bmatrix}_i \right\| \tag{13}$$

where $\|.\|$ denotes the L2-norm function. An important characteristic of Algorithm 3 is that the best subset is determined only based on the properties of the affine transformation as well as the systematically derived ellipse validation criterion without requiring any subjectively defined and scene specific thresholds.

The overall robust ellipse detection algorithm can be summarized as follows:

1- Define $N$ random subsamples with $k$ samples using Algorithm 1.
2- For each subsample, $i = 1 \ldots N$, perform Algorithm 2 to obtain, $in_e$ and $(r_s, x_s, y_s)_i$.
3- Perform Algorithm 3 to determine the best inlier set amongst all subsamples.

Even though the proposed robust ellipse fitting can be applied to all connected segments, to preserve computational efficiency, as described in the beginning of Section 3, it will only be applied to: (i) connected segments whose best fit ellipse does not intersect with an existing validated ellipse (Figure 3f); and (ii) two or more validated ellipses whose best fit ellipses intersect (Figure 3g).

*3.6. Post-processing for 3D point clouds*

The geometric parameters of ellipses, detected from 2D projected points, can be used to derive an initial estimate of the geometric parameters of the original cylinder as follows (Miller and Goldman, 1992):

$$\begin{cases} R = b_e \\ \delta = \cos^{-1} \frac{b_e}{a_e} \\ C_c = Rot^T . C_e \end{cases} \tag{14}$$

where $R$ and $C_c$ are the radius, and the point of intersection between the cylinder's axis and the plane, respectively, $C_e = (x_e, y_e, z_0)^T$, $a_e$ and $b_e$ are the center, semi-major length and semi-minor length of the ellipse in the coordinate system of the intersecting plane, respectively, $Rot$ is the rotation matrix that takes the plane's normal vector to vector $(0,0,1)^T$, and $\delta$ is the angle between the plane's normal vector and the cylinder's axis. To acquire the cylinder's axis from equation 14, the normal vector must be rotated $+\delta$ or $-\delta$ along the semi-minor axis (i.e. along the vector $Rot^T.(-\sin(\theta),\cos(\theta),0)^T$)[6]. Therefore, two possible solutions exist. To determine which axis is the correct one, **Algorithm 4: Cylinder Detection from Ellipses**, in combination with the robust cylinder detection of (Maalek et al., 2019), is developed as follows:

1- From the geometric parameters of the ellipses, estimate $\delta, R, C_c$ using equation 14.
2- Determine the two possible cylinder axes by rotating the plane's normal $+\delta$ or $-\delta$ along the ellipses' minor axis.
3- For each of the two estimated cylinder axes, perform the following:
    a) Rotate the points such that the cylinder's axis is parallel to the z-axis, using rotation matrix, $Rot_{az}$.
    b) Find all points within $\gamma R$ from the rotated center, $Rot_{az}.C_c$, in the x-y plane.
    c) Perform the robust circle fitting using Algorithms 1 and 2 of (Maalek et al., 2019).
4- Find the cylinder's axis corresponding to the robust circle fit that achieved the lowest root mean squared error (RMSE)-among the two candidates. This axis is considered as the initial cylinder's axis.
5- Perform Algorithm 3 of (Maalek et al., 2019), robust cylinder axis estimation, on the rotated points, corresponding to the initial estimated cylinder's axis. This algorithm provides the final geometric parameters of the cylinder along with the points following the pattern of the best fit cylinder.

The last consideration for Algorithm 4 is to define the radius, $\gamma R$. This radius must be large enough to ideally encompass all points belonging to the cylinder. Here, following (Maalek et al., 2019), the cylindrical radius is defined as: $\gamma R = \max(2R, R + 10cm)$. For our datasets, this radius was sufficient to completely encompass the desired cylinder.

*3.7. Metrics for validation of proposed methods*

The designed experiments mainly involve the comparison of ellipse extraction quality using different methods. Here, the four main metrics, commonly used to determine the quality of object extraction algorithms, namely, precision, recall, accuracy, and F-measure (also known as F1-score), are utilized. These metrics are estimated as follows:

$$\begin{cases} Precision &= \frac{T_P}{T_P+F_P} \\ Recall &= \frac{T_P}{T_P+F_N} \\ Accuracy &= \frac{T_P+T_N}{T_P+T_N+F_P+F_N} \\ F-measure &= 2.\frac{Precision.Recall}{Precision+Recall} \end{cases} \qquad (15)$$

where $T_P, T_N, F_P, F_N$ are the number of true positive, true negative, false positive, and false negative of the detection, respectively. To calculate the accuracy of the estimated parameters of best fit cylinders and ellipses, the L2-norm of the estimated parameters from the final ground truth parameters were used.

## 4. Experimental Design

For this work, four classes of experiments were designed, namely comparative analysis of robust ellipse fitting, sensitivity analysis of ellipse validation criterion, comparison of non-overlapping ellipse detection from images, and cylinder extraction from point clouds. They are explained in detail in the following subsections[7].

---

[6] Based on Rodrigues' formula, the exact equations for the rotation matrices were presented in eq. 4 of (Bergamasco et al., 2020).
[7] Computations were carried out using a desktop with Ryzen 5-2600X CPU, 64 GB RAM, and 1TB SSD NVME storage.

*4.1. Comparative analysis of robust ellipse fitting*

Four experiments were designed in this category to assess the impact of the proposed sampling as well as the overall robust ellipse fitting method:

1- **Importance of sampling strategy:** this experiment evaluated the performance of the proposed sampling strategy in comparison to random sampling, on simulated ellipses with no outliers. To this end, a base ellipse configuration with rotation angle, aspect ratio[8], semi-minor length, noise, and number of points of $\frac{\pi}{4}$, 2, 10, 0.5, 50, respectively, was considered (simulated using Algorithm 2 of (Maalek and Lichti, 2020)). Two sets with 10,000 subsets of size $k = 6$, using both our method (Algorithm 1) and random sampling, were generated. For each subset, the best fit ellipse was fitted to the sets of six points and the L2-norm of the estimated geometric parameters to that of the actual was estimated. To account for the impact of random measurement errors, each subsample was selected from 500 simulated ellipses, and the average of the parameter error for each subsample was reported.

2- **Impact of outlier ratio on robust ellipse detection:** this experiment investigated the effects of increasing the outlier ratio on the propose robust ellipse fitting method, compared to LMedS (described in Section 2.1), spatial averaging of random ensembles (SARE) (Thurnhofer-Hemsi et al., 2020) and Munoz' (Muñoz-Pérez et al., 2014) method. LMedS method was chosen as a basis for comparison since it is an established method for robust ellipse fitting that alike ours does not require subjective tuning of parameter thresholds. The other two methods are also more recent robust methods, which were used to fit ellipses in contaminated points with promising results. To this end, a base ellipse configuration, comprised of rotation angle, aspect ratio, semi-minor length, noise, and number of points equal to $\frac{\pi}{4}$, 2, 1, 0.01, 1000, respectively was simulated. Outlier points following the patterns of two larger rectangles were added (see Figure 10b). The aspect ratio, center, rotation angle, and noise of the rectangles were the same as the ellipse. The width and length of the two rectangles were chosen as 1.5 and 2 units larger than the minor and major lengths of the ellipse, respectively[9]. The outlier ratio was increased from 0.25 to 0.75 in 0.05 increments. In each outlier instance, the number of points, representing the outliers were shared evenly between the two rectangles. For each outlier ratio, the F-measures using our method, LMedS with our proposed sampling (Algorithm 1), LMedS with random sampling, LMedS, SARE, and Munoz were reported.

3- **Impact of noise on robust ellipse detection:** following the previous experiment, this experiment instead investigated the impact of noise. Therefore, the normalized noise (noise per square root of the area of the ellipse) was changed from 0 to 0.05 in increments of 0.0025 units. All other simulated configurations are identical to the previous experiment, except for the outlier ratio, which was chosen as 0.5. The performances of our method, LMedS with our sampling, LMedS with random sampling, SARE, and Munoz were evaluated as a function of the change in the normalized noise.

4- **Robust ellipse detection comparison:** this experiment evaluated the quality of the robust ellipse extraction using our proposed method, compared to the LMedS method. To this end, 10 different sets of connected points, comprised of ellipses and outlier points, were chosen from the edge points generated from real-world images as well as projected point clouds of cylinders. The quality of ellipse point extraction, i.e. precision, recall, accuracy, and F-measure, of our method, LMedS with random sampling, and LMedS with our proposed sampling were measured[10]. For all methods, the sample size and the number of subsamples were chosen as $k = 6$ and $N = 1,500$, respectively. 1,500 subsamples are roughly equivalent to finding at least one outlier free subsample with 99.5% probability from data with an outlier ratio of, 0.6.

---

[8] The ratio of the semi-major length to the semi-minor length of an ellipse
[9] Larger to promote minimal overlap between ellipse and rectangular points as the noise increases.
[10] The ground truth elliptical points were determined manually.

*4.2. Sensitivity analysis of ellipse validation criterion*

The following three experiments were designed to investigate the sensitivity of the proposed ellipse validation criterion in both simulated and real-world datasets:

1- **Comparison of ellipticity measures:** this experiment was designed to empirically determine the method and threshold required to differentiate between ellipses and other ellipse-like objects (i.e. rectangles) in a connected set of points. To this end, 10,000 ellipses and rectangles were randomly generated, subjected to the following configurations:
   a) **rotation angle,** randomly chosen between 0 and $\pi$;
   b) **aspect ratio,** randomly chosen between 1 and 4;
   c) **measurement standard deviation,** randomly chosen between 0 and 5 units;
   d) **semi-minor length,** randomly chosen between 10 and 100 units;
   e) **number of points,** randomly chosen between 50 and 400; and
   f) **ellipse spanning arc,** randomly chosen from either full, three quarter, or half ellipse.
   To provide a fair comparison, the parameters are shared between the ellipse and rectangle at each configuration. For each generated ellipse and rectangle, all three ellipticity measures described in Section 2.2, namely, elliptic variance ($E_V$; equation 4), moment invariants ($E_I$; equation 5), and Euclidian ellipticity ($E_E$; equation 6), were calculated and compared.

2- **Impact of noise on the Euclidian ellipticity:** the purpose of this experiment was to determine the minimum Euclidian ellipticity that can effectively differentiate between ellipses and rectangles as the normalized noise (measurement error per square root of ellipse area) increases. To this end, a base configuration with the rotation angle, aspect ratio, semi-minor length, and number of points, chosen as $\frac{\pi}{4}$, 2, 10, 50, respectively, is defined. The normalized noise is then changed from 0 to 0.1, and the Euclidian ellipticity is measured for both rectangles and ellipses. To effectively capture the behavior of random measurement errors, each parameter configuration was simulated 500 times and the mean, minimum and maximum of the Euclidian ellipticity for both ellipses and rectangles were reported.

3- **Sensitivity analysis on real image edge points:** this experiment investigated the sensitivity of the ellipse detection quality (i.e. precision, recall, accuracy, and F-measure) as the Euclidian ellipticity changes from 0.9 to 1. The real-world data is acquired from a calibration laboratory (Figure 1a). A total of thirty images with different orientations, captured via iPhone 11, were used in this study.

*4.3. Comparison of non-overlapping ellipse detection from images*

This experiment evaluated the quality of the proposed ellipse detection method, compared to two established edge-based methods, presented in Fornaciari (Fornaciari et al., 2014) and Patraucean (Pătrăucean et al., 2017), and a shape-based method, proposed in Panagiotakis (Panagiotakis and Argyros, 2020). The method of Fornaciari was selected to represent a typical edge chaining method with reported reliable results. The method of Patraucean was chosen to provide a representation of the state-of-the-art parameter-less methods for ellipse validation, which also reported reliable results. Finally, the method of Panagiotakis was used as a parameter-less shape-based method, which is suggested to be more tolerant to noise than edge-based methods, with promising and reliable results. The thirty images, used in the previous experiment, were employed to report, and compare the precision, recall, accuracy, and methods.

### 4.4. Cylinder detection from point clouds

In this experiment, two sets of point cloud data, one from a controlled laboratory environment, and another from a real-world industrial construction site, were used. The two datasets provide baselines on the applicability of our proposed cylinder detection method, in both a controlled laboratory setup as well as a real industrial plant under construction with many occlusions. The laboratory data was captured from multiple professionally installed mechanical pipes in the calibration laboratory at the University of Calgary, using the Leica HDS6100 terrestrial laser scanner (TLS). The real-world point cloud was acquired from a portion of an industrial plant, using a FARO Focus TLS, with multiple pipes with different orientations. The complete dataset of the latter construction site is publicly made available by ClearEdge3D (ClearEdge3D, 2020). For each point cloud dataset, the following steps were followed:

1. Three planes that intersected with the cylinders were manually selected.
2. For each plane, all points within 1mm of each plane were projected.
3. For each projection, the proposed non-overlapping ellipse detection method was carried out to detect the available ellipses.
4. The cylinders were then detected using Algorithm 4.
5. The accuracy of the estimated cylinder parameters using only ellipse fitting (equation 14) compared to the ground truth (final detected cylinder using the method of (Maalek et al., 2019)) was reported.

### 4.5. Summary of experiments

Table 1 shows a summary of the experiments carried out in this manuscript along with the type of data and the experiment's purpose towards validating the original methods presented in Section 3. From the summarized table, it can be observed that three types of data are used during the experiments, namely simulated ellipses and rectangles, images from a calibration target field, and point clouds of scenes with cylindrical pipes. The ellipses and rectangles are simulated using the method proposed in (Maalek and Lichti, 2020). The images are acquired using an iPhone 11. The point cloud datasets contained multiple pipes under controlled laboratory and sub-optimal industrial construction site conditions. For the reader's reference, Figure 7 provides sample images of the data used in this study for different experiments.

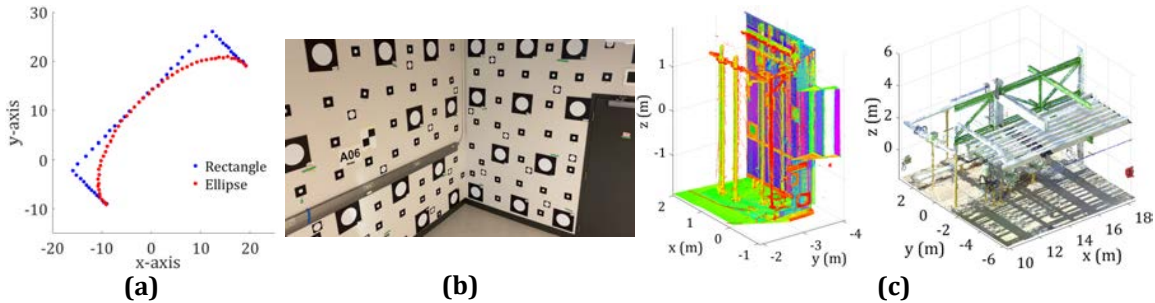

**Figure 7:** Sample data used in this manuscript: a) simulated ellipse (red), and rectangle (blue); b) images from the circular calibration target field; and c) the collected TLS point clouds of professionally installed mechanical pipes (left), and industrial plant (right)

Table 1: summary of the designed experiments

| Experiment Description | | Type of data | Purpose |
|---|---|---|---|
| **Comparative analysis of robust ellipse fitting** | Importance of sampling strategy | Simulated ellipses | Comparing Algorithm 1 to random sampling |
| | Impact of outlier ratio on robust ellipse detection | Simulated ellipses and rectangles | Comparing Algorithms 1-3 to LMedS, SARE (Thurnhofer-Hemsi et al., 2020) and Munoz (Muñoz-Pérez et al., 2014) as a function of outlier ratio |
| | Impact of noise ratio on robust ellipse detection | Simulated ellipses and rectangles | Comparing Algorithms 1-3 to LMedS, SARE and Munoz as a function of normalized noise |
| | Robust ellipse detection comparison | Real-world image edge points and projected point clouds | Comparing Algorithms 1-3 to LMedS |
| **Sensitivity analysis of ellipse validation criterion** | Comparison of ellipticity measures | Simulated ellipses and rectangles | Comparing Euclidian ellipticity to elliptic variance and moment invariants |
| | Impact of noise on the Euclidian ellipticity | Simulated ellipses and rectangles | Determining a lower bound for the Euclidian ellipticity |
| | Sensitivity analysis on real image edge points | Real-world images | Evaluating the ellipse detection quality as a function of ellipticity threshold |
| **Comparison of non-overlapping ellipse detection from images** | | Real-world images | Comparing our ellipse detection quality, to the methods of (Fornaciari et al., 2014), (Pătrăucean et al., 2017), and (Panagiotakis and Argyros, 2020) |
| **Cylinder detection from point clouds** | | Two sets of real-world TLS point clouds | Application of non-overlapping ellipse detection to automate cylinder extraction |

## 5. Experimental results

*5.1. Comparative analysis of robust ellipse fitting*

*5.1.1. Importance of sampling strategy*

To provide some perspective on the importance of the proposed sampling strategy, the accuracy of the fitted ellipses in 10,000 subsets of six points, selected randomly and using our sampling strategy (Algorithm 1), were compared. Figure 8 shows the results of the parameter errors obtained from the two methods. As can be observed, the proposed sampling strategy provides significantly more reliable estimations of the ellipse parameters for all subsets since it eliminates the possibility of sampling adjacent points. On the other hand, the random sampling produced up to 300 times poorer best fit ellipse parameter estimations since there are no restrictions on the relative geometry of points in each subset. In fact, about 62% of the randomly-acquired subsets obtained parameter errors higher than the worst subset using our sampling method. These subsets are practically useless as they are unable to predict the true pattern of the best fit ellipse to a set of data. Therefore, the method of sample selection plays a crucial even when dealing with data only subjected to random measurement errors with no outliers.

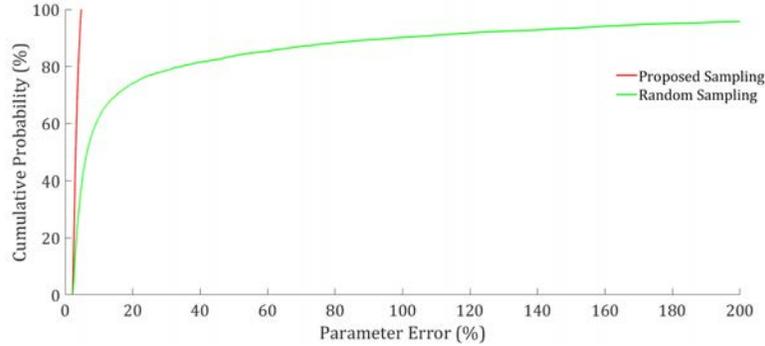

**Figure 8:** Mean of the best fit ellipse parameters obtained by 10,000 subsets of six points using our sampling method (red), and random sampling (green)

*5.1.2. Impact of outlier ratio on robust ellipse detection*

    The impact of ratio of outliers on the quality of the robust ellipse detection using our method (Algorithms 1-3), compared to LMedS, SARE (Thurnhofer-Hemsi et al., 2020), and Munoz (Muñoz-Pérez et al., 2014), was determined. To this end, the outlier ratio was increased from 25% up to 75% and the F-measure achieved using our method, LMedS with our sampling strategy, LMedS with random sampling, SARE (Thurnhofer-Hemsi et al., 2020), and Munoz (Muñoz-Pérez et al., 2014) were calculated. Figure 9a demonstrates the results of the F-measure as a function of the outlier ratio. The following note-worthy observations were made. First, our method and LMedS (regardless of sampling strategy) outperformed the methods of SARE, and Munoz for all outlier ratios. SARE and Munos appear to be impacted by the outlier ratio in a close to linear manner. Secondly, for outlier ratios between 25% and 50%, our method, LMedS with our sampling strategy, and LMedS with random sampling performed very similarly and correctly detected the elliptic points from the outliers with very high F-measure. As the outlier ratio increased from 0.5 to 0.75, however, LMedS using both sampling strategies started detecting non-elliptic points as elliptical, which increased Type II errors, and consequentially decreased the F-measure. The outlier ratio of 0.5 is just above the breakdown point of LMedS, and hence, the method was expected to become less effective as the outlier ratio increased. Our method, on the other hand, is not impacted by the outlier ratio and consistently produced reliable results. Figures 9b-9d provide some visual perspective on the detected elliptic points using the different methods, applied to data with the outlier ratio of 0.25, 0.5 and 0.75, respectively. As observed, our method extracted the elliptic points correctly, regardless of the increase in the number of outliers, and outperformed the considered methods.

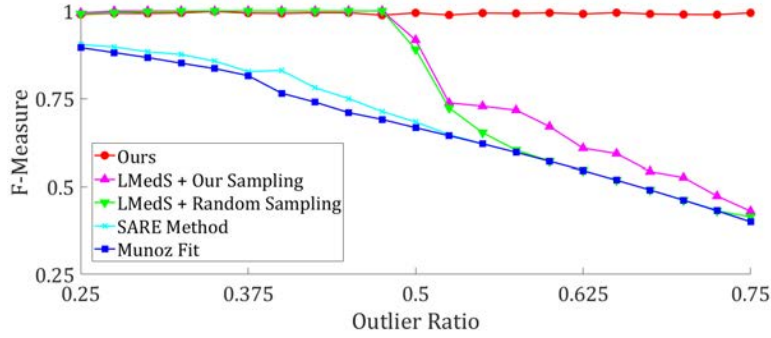

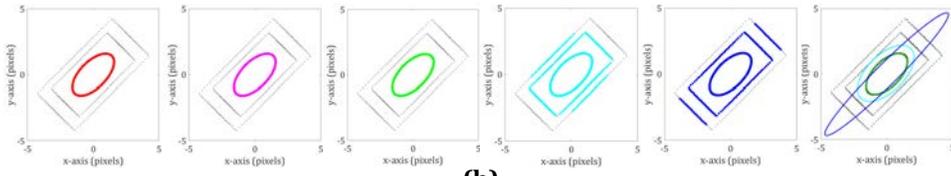

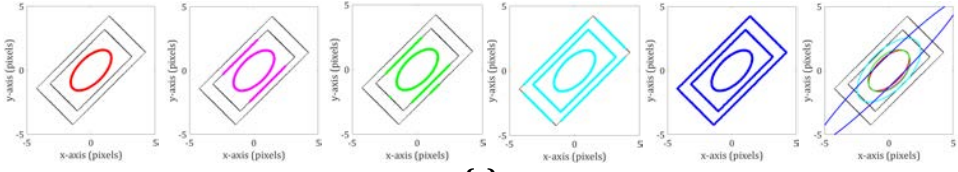

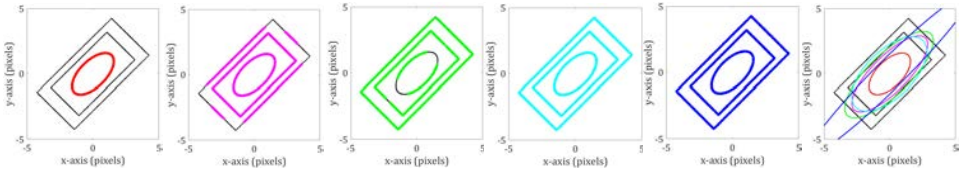

**Figure 9:** a) Impact of outlier ratio on the F-measure, achieved using our method (red), LMedS with our sampling (magenta), LMedS with random sampling (green), SARE (cyan), and Munoz (blue). Detected elliptic points and best fit ellipses (far right) using different methods as a function of outlier ratio: b) 0.25; c) 0.5; and d) 0.75

### 5.1.3. Impact of noise on robust ellipse detection

The impact of noise was evaluated on the quality of the ellipse detection using our method, LMedS, SARE and Munoz. The noise per unit area was changed from 0 to 0.05 in 0.0025 increments. The outlier ratio was chosen as 0.5. The F-measure using our method, LMedS with our sampling, LMedS with random sampling, SARE and Munoz as a function of the normalized noise is shown in Figure 10a. It was observed that SARE and Munoz were consistently less reliable than LMedS and our method for all noise levels. It was also observed that the LMedS using our sampling strategy slightly outperformed LMedS with random sampling. Overall, our method achieved considerably more reliable results compared to the other four methods. The LMedS method showed a small decline in the F-measure as the noise increased. Our method showed relatively stable results up until the normalized noise of 0.05. To provide a visual perspective, the simulated points with normalized noise of 0.01, 0.025 and 0.05 are shown in Figures 10b, 10c, and 10d, respectively. As observed, our method detected the elliptic points effectively as the noise level increased. The LMedS, SARE and Munoz methods, on the other hand, identified more outliers as elliptical, as the noise level increased. From Figures 10b to 10d, it can also be observed that the boundary between the ellipse and outer rectangle becomes less detectable as the noise level increases; however, our method was still able to differentiate the ellipse from the rectangles accurately.

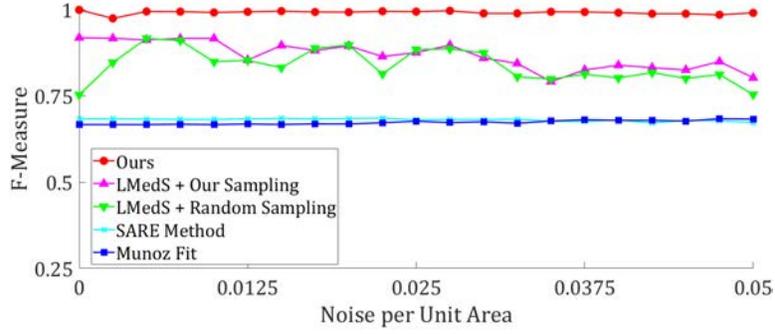

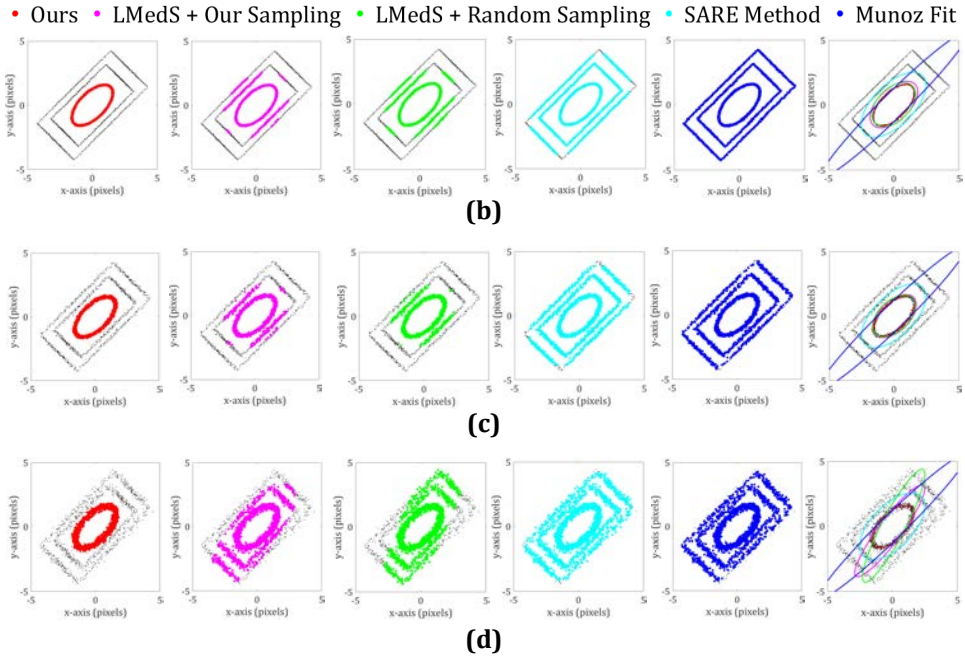

**Figure 10:** a) Impact of noise per unit area of ellipse on the F-measure, achieved using our method (red), LMedS with our sampling (magenta), LMedS with random sampling (green), SARE (cyan), and Munoz (blue). Detected elliptic points and best fit ellipses (far right) using different methods as a function of noise per unit area: b) 0.01; c) 0.025; and d) 0.05

*5.1.4. Robust ellipse detection comparison*

The quality of the proposed ellipse extraction method using Algorithms 1-3, compared to the LMedS method, was evaluated on real-world datasets. Ten samples of connected points, containing a combination of ellipse and outliers, were chosen from both edge points of real-world images, and projected points of real-point clouds. The data on average contained 58.6% outliers, suggesting that on average 41.4% of the points forming an elliptical pattern required to be extracted. Figures 11 and 12 show the extracted elliptic points using our method (red), LMedS with our proposed sampling strategy (magenta), and LMedS with random sampling (green). Similar to the simulation results, our proposed method showed promising capabilities in extracting the elliptic points from the outliers in all ten cases, whereas the extracted points using the LMedS method, regardless of the sampling strategy, contained many Type I and Type II errors. The overall precision, recall, accuracy, and F-measure for each method are reported in Table 2. As observed, our method significantly outperformed the LMedS method regardless of sampling strategy. The LMedS method using our sampling strategy, however, provided slightly better overall extraction results compared to purely random sampling. Therefore, even though the importance of sampling strategy was demonstrated, it must be paired with an appropriate method to produce reliable and desirable overall ellipse extraction results. The results obtained by both simulated and real-world experiments suggest that the propose robust ellipse detection method outperforms that of LMedS.

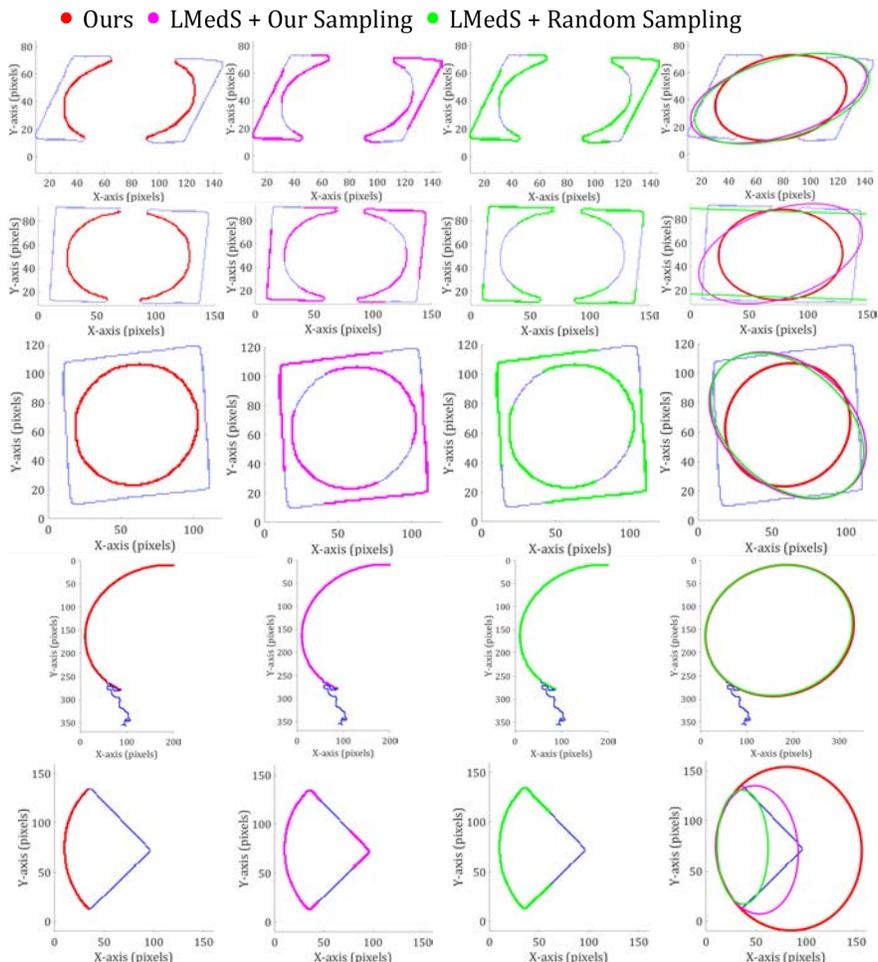

**Figure 11:** Robust extraction of points following elliptical patterns for five samples, acquired from real-world image edge points: left- our proposed method (red); middle left- LMedS with our proposed sampling strategy (magenta); middle right- LMedS with random sampling (green); and right- best fit ellipse to the set of points using all three methods.

**Table 2:** Precision, recall, accuracy, and F-measure (F1-score) for ellipse extraction using our, LMedS with our sampling strategy and LMedS with random sampling

| Method | Precision | Recall | Accuracy | F-Measure |
|---|---|---|---|---|
| Ours | 97.3% | 98.8% | 98.4% | 98.1% |
| LMedS- our sampling | 83.9% | 46.8% | 57.5% | 60.1% |
| LMedS- random sampling | 82.8% | 43.7% | 53.7% | 57.2% |

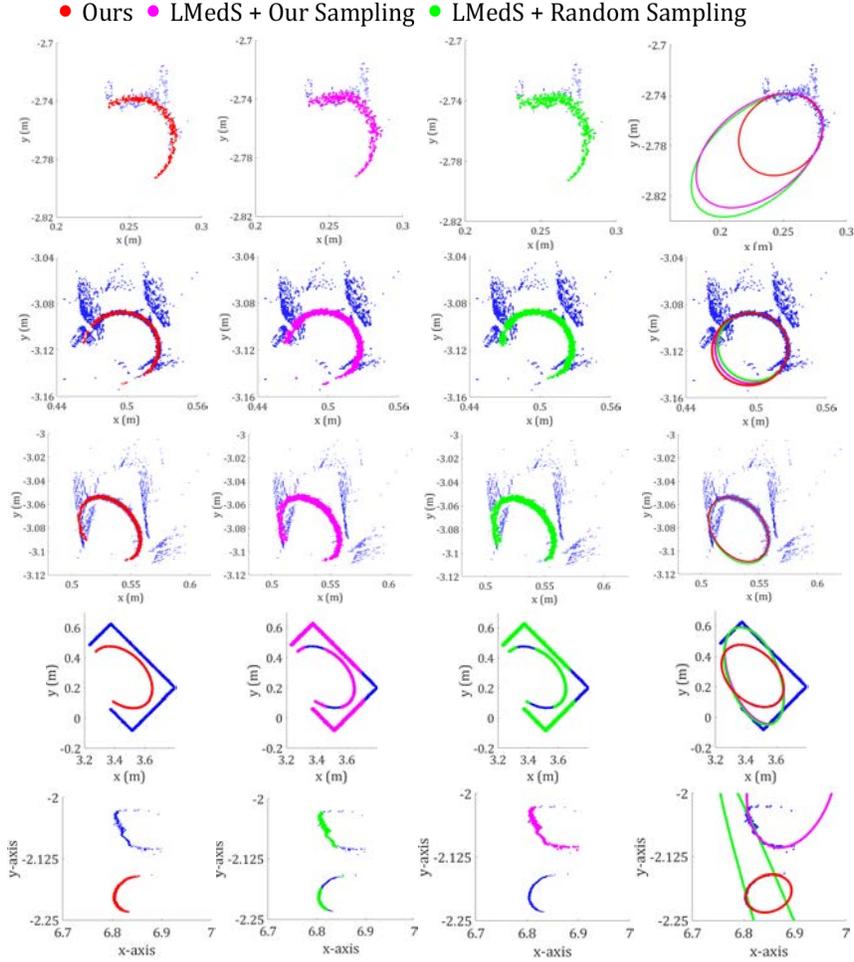

**Figure 12:** Robust extraction of points following elliptical patterns for five samples, acquired from real-world projected point clouds: fat left- our proposed method (red); middle left- LMedS with our proposed sampling strategy (magenta); middle right- LMedS with random sampling (green); and far right- best fit ellipse to the set of points using all three methods.

## 5.2. Sensitivity analysis of ellipse validation criterion

### 5.2.1. Comparison of ellipticity measures

As described in Section 4.2, 10,000 different ellipses and rectangles were generated to evaluate the behavior of three ellipticity measures, namely, Euclidian ellipticity, ellipticity variance, and moment invariants, introduced in Section 2.2. The result of the simulation is presented in Figure 13. For each chart of Figure 13, the horizontal axis represents the value of the specific ellipticity measure, the vertical axis represents the cumulative probability, and the blue and red curves represent the results for rectangles and ellipses, respectively. As illustrated, elliptic variance (Figure 13b) and moment invariants (Figure 13c) do not show a specific gap between the curve representing the ellipse and that of the rectangle. This suggests that a threshold, above which all (or most) ellipses lie and no (or a very few) rectangles exist, cannot be determined. For instance, there is over 60% and 98% chance of rectangles and ellipses obtaining a moment invariant of above 0.8, respectively. This suggests that if the goal is to detect ellipses with 98% confidence (i.e. moment invariant of above 0.8) using moment invariants, the results will also include more than a 60% chance of classifying rectangles as ellipses (significant Type II error for detection). The same analogy can also be used for elliptic variance. Euclidian ellipticity (Figure 13a), on the other hand, provides a clear gap between the rectangles and ellipses. In fact, as shown in the green dashed line, out of the 10,000 different ellipses and rectangles, 100% and 0% of ellipses and rectangles, respectively obtained Euclidian ellipticity of larger than 0.96. In other words, it is expected that all ellipses fall above, and all rectangles fall below the Euclidian ellipticity of 0.96. It is important to mention that if the Euclidian ellipticity is selected above 0.965, some ellipses might not be classified (Type I error) even though no rectangles will be classified as ellipses (robustness to Type II errors). Therefore, the Euclidian ellipticity of 0.96 appears to provide robustness to both Type I and Type II errors, which was used in this study as described in Section 3.4 - equation 7.

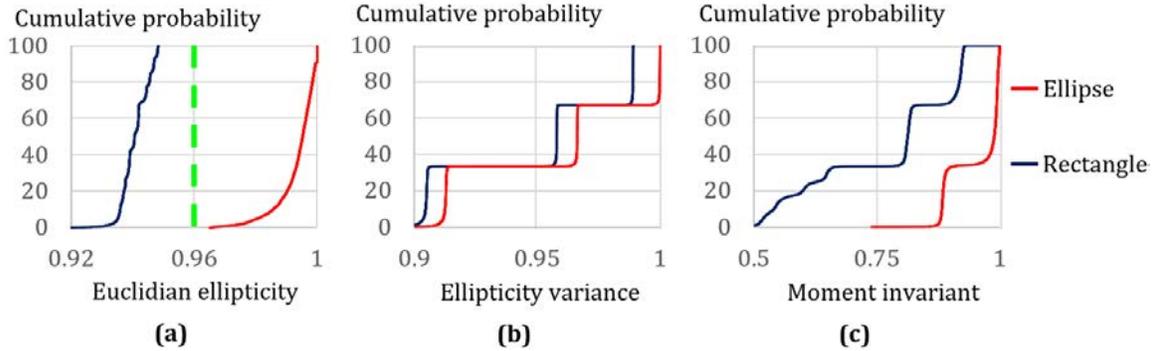

**Figure 13:** Results ellipticity measures of 10,000 simulated ellipse and rectangles: a) Euclidian ellipticity; b) elliptic variance; and c) moment invariants

*5.2.2. Impact of noise on the Euclidian ellipticity*

This section provides further analysis on the impact of measurement standard deviation normalized by the area of an ellipse on the Euclidian ellipticity. To this end, the normalized standard deviation is increased from 0 to 0.1 in 0.005 increments, and the Euclidian ellipticity for ellipses and rectangles with the base configurations, presented in Section 4.2, were calculated. Each configuration was simulated 500 times. Figure 14 shows the results of the mean, minimum and maximum Euclidian ellipticity for ellipses and rectangles as the normalized noise increases. It is observed that as the normalized noise increases, the mean of the Euclidian ellipticity for ellipses and rectangles start approaching each other. In addition, the confidence regions (the area confined between the minimum and maximum) for both ellipses and rectangles increase as the noise increases. This shows that additional noise increases the uncertainty in the Euclidian ellipticity estimation, which in turn creates obstacles in differentiating between ellipses and rectangles. Therefore, there is a lower limit for the Euclidian ellipticity because of the increase in the normalized noise levels. Figure 14 shows the Euclidian ellipticity, below which ellipses and rectangles may not be effectively differentiated. In our simulations, the maximum Euclidian ellipticity achieved by all rectangles was around 0.955. Hence, the threshold, $Th_E$, must be larger than 0.955 to account for Type II errors in the ellipse detection (i.e. no rectangles classified as ellipses).

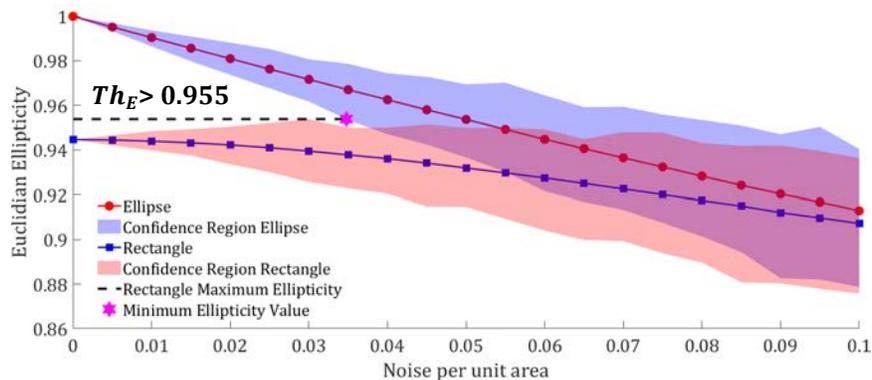

**Figure 14:** Impact of the normalized noise on the Euclidian ellipticity for both simulated rectangles and ellipses

*5.2.3. Sensitivity analysis on real image edge points*

From the simulated results of Sections 5.2.1 and 5.2.2, the lower and upper bounds for the Euclidian ellipticity were determined as 0.955 and 0.965, respectively. The lower and upper bounds are required to minimize Type II and Type I errors in the detection of ellipses, respectively. To identify the effectiveness of the upper and lower bounds determined through simulated experiments, the proposed non-overlapping ellipse detection method was applied to thirty images from the calibration laboratory of Figure 1a. The overall precision, recall, accuracy, and F-measure for all thirty images combined were calculated as the $Th_E$ of equation 7 was changed from 0.9 to 1 in 0.025 increments. Figure 15a shows the results of the precision, recall, accuracy, and F-measure as a function of the Euclidian ellipticity threshold. It can be observed that for $Th_E$ between 0.9 and 0.925, the quality of the extraction remains relatively unchanged. As suspected, all elliptical segments were detected as ellipses (low Type I error and high precision), and many other segments were also classified as ellipses (high Type II errors shown by the lower recall rate). As the threshold increases from 0.9425, the precision remains unchanged while the recall improves (which indicates lower Type II errors). The optimum solutions for precision, recall, accuracy, and F-measure, were obtained when the $Th_E$ was between 0.955 and 0.965, which were the upper and lower bounds suggested from the simulated experiments. As the $Th_E$ increases from 0.965 to 1, lower number of elliptic segments in the data were classified as ellipses (Type I errors increase, which contributes to the decrease in precision).

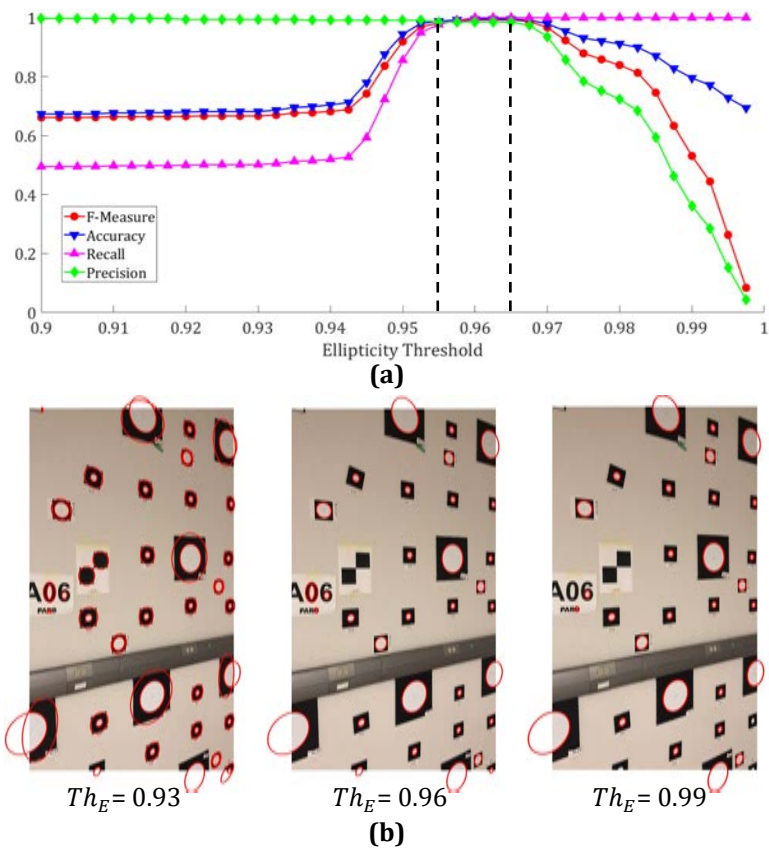

**Figure 15:** Impact of Euclidian ellipticity threshold on: a) precision, recall, accuracy, and F-measure of ellipse detection from thirty images; and b) extracted ellipses for the sample image

To provide some perspective, the ellipses extracted from the image of Figure 1a using three ellipticity thresholds of 0.93, 0.96 and 0.99 are shown in Figure 15b. Using thresholds 0.93, 0.96 and 0.99, the number of detected ellipses were 66, 34 and 29, respectively. $Th_E = 0.93$ included many non-elliptical shapes such as rectangles as ellipses (Type II errors). $Th_E$=0.96 was able to detect all ellipses correctly without classifying any non-elliptic segments (robust to both Type I and Type II errors). $Th_E = 0.99$ achieved robustness to Type II errors but did not classify 5 out of the 34 ellipses, which consequentially negatively impacts the precision of the detection. Predicated on the results of the simulated and real-world experiments, the $Th_E = 0.96$ can be recommended, which produced a reliable balance between the Type I and Type II errors for the problem of ellipse detection from 2D points.

*5.3. Comparison of non-overlapping ellipse detection from images*

The overall quality of the proposed method for detecting non-overlapping ellipses from images is evaluated in this section. Figure 16 shows six sample images (from the thirty used in this experiment) taken from a corner of a complete calibration laboratory. The images of Figures 16a, 16b, 16c, 16d, 16e and 16f represent cameras located at average distances of approximately 1.25m, 1.5m, 1.75m, 2m, 2.5m, and 3.5m from the targets, respectively. Three different types of ellipse detections are presented in Figure 16. The first type of ellipse colored as magenta, which contain most of the detected ellipses in all images, belong to segments with no outliers. These segments only contain the edge points representing one ellipse, and their best fit ellipse satisfies the condition imposed by equation 7 (detected ellipses of Figure 3e). The second type of ellipse, colored as green, belong to segments where connected edge points of one ellipse also contain outliers due to errors in pre-processing, such as binarization errors (see Figure 3f). These ellipses are detected using a single implementation of the proposed robust ellipse fitting (Algorithms 1-3). The last category (cyan color) are ellipses that not only contain outliers (similar to the last category), but the points representing a single ellipse are also divided into two or more segments (see Figure 3g). These ellipses are detected by first applying the robust ellipse detection on each segment individually to determine the points following elliptic patters. Since these elliptic points are from the same ellipse, their best fit ellipses will overlap. These overlapping segments are then grouped into one, and the robust ellipse fitting is again applied on the unified segment to determine the final set of elliptic points along with the best fit ellipse.

Figure 16a shows the ellipses detected from an image of targets attached to one planar wall. Since the image was taken relatively close with little to no visible local variations, all ellipses of targets were confined in one segment (ellipses of the first category), which were then differentiated from the remaining segments using the validation condition of equation 7. Figure 16b shows an image where one of the segments, due to errors during binarization, was from the second category, which was then correctly extracted using our robust method. Figures 16c and 16d shows situations where a slight shake in the camera during image acquisition—and the consequential errors produced during binarization—had divided multiple ellipses of different targets into two or more segments, which required the robust ellipse fitting twice (third ellipse category). Similarly, Figure 16e shows a visible shake during image acquisition, which had created distortions. As observed, our method is still capable of detecting the ellipses without additional subjective fine tuning of parameters during the pre-processing stage. Figure 16f shows an image taken from a relatively farther location to the targets at quarter of the resolution compared to the other images (1080p instead of 4K). As observed, our method detected even the ellipses of the relatively small targets (semi-minor of around 3-4 pixels) in all cases. To provide additional perspective on the relative performance of our method compared to other state-of-the-art edge methods, the real-world images were also processed using the ellipse detection methods of (Fornaciari et al., 2014), (Pătrăucean et al., 2017), and (Panagiotakis and Argyros, 2020). The results of the detected ellipses using the methods of Fornaciari, Patraucean and Panagiotakis, applied to four out of the six samples presented in Figure 16, are shown in Figures 17, 18, and 19, respectively.

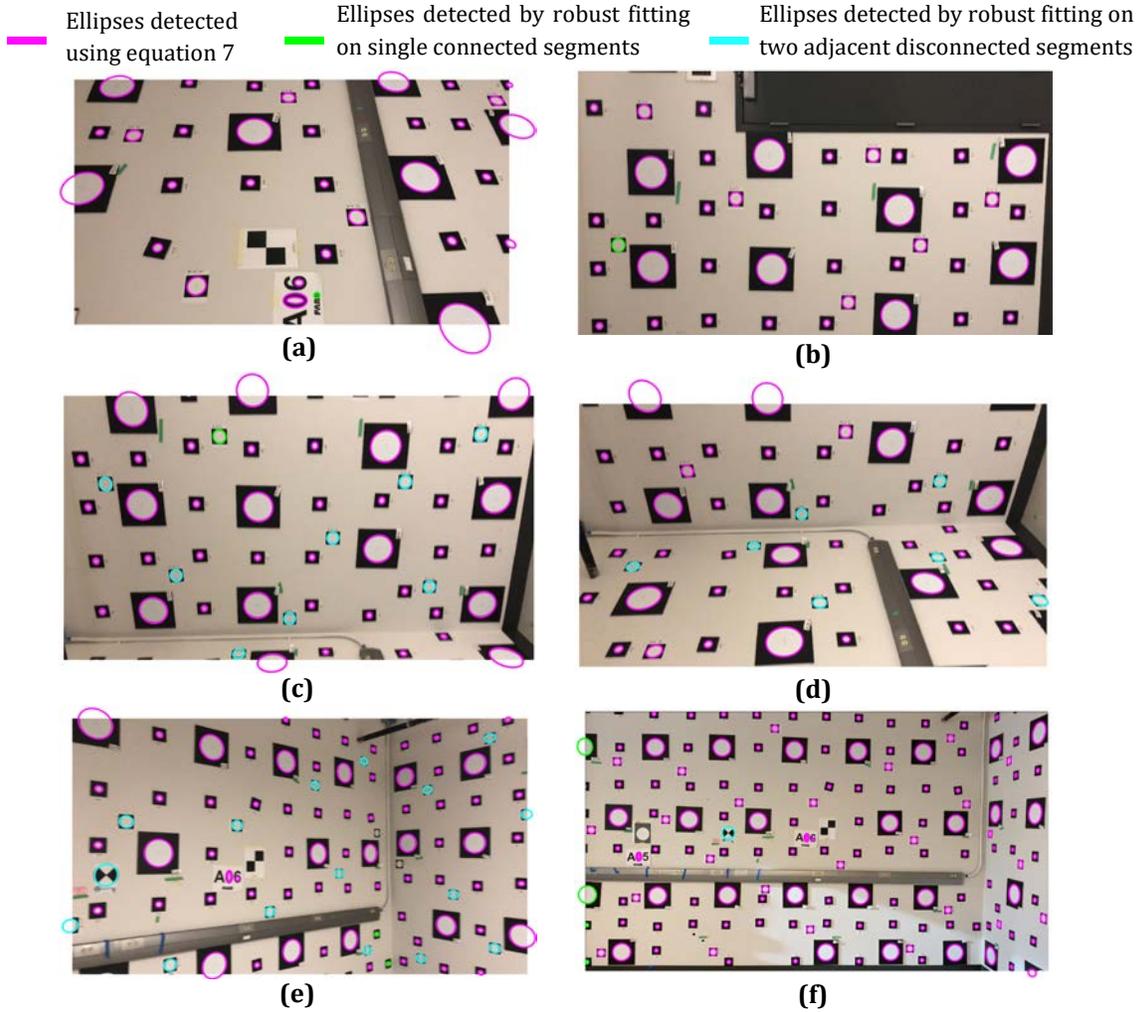

**Figure 16:** Types of ellipses detected using our proposed method: ellipses with no outliers (magenta); ellipses with outliers in one segment (green; the case of Figure 3f); and ellipses with outliers divided into two or more segments (cyan; the case of Figure 3g). a) image where edge points of targets representing ellipses contained no outliers; b) image where edge points representing an ellipse contained outliers in a single connected segment (green); c) image where all three types of ellipses were detected (magenta, green and cyan); d) image where edge points representing a few ellipses contained outliers and were divided into multiple segments (cyan); e) image with visible distortions containing all three types of ellipses; f) image taken farther away (compared to the rest) containing all three types of ellipses.

From the visual assessment of Figures 17 through 19, the following observations were made:

1- All three methods, especially for the images of Figures 16e and 16f, were unable to detect smaller ellipses, which increases Type I errors, and contributes directly to lower extraction precision.
2- Non-elliptic edges were detected as elliptic in many instances (see Figures 17e, 18e, and 19e), which increases Type II errors, and consequentially lowers the recall rate in both methods.
3- Multiple ellipses were fitted to a single connected segment on many occasions (Figures 17f, 18f and 19f). This shows that their proposed strategies to merge overlapping ellipses of the same arc must be revisited.
4- The fitted ellipse parameters to a connected elliptic segment were incorrect and biased in many cases (Figures 17g, 18g and 19g). This is attributed to the fact that in these examples only a portion of an elliptic arc was detected as elliptical (the remaining points of the same ellipse were detected as non-elliptical), and as discussed, ellipses fitted to small fractions of an ellipse can be biased and inaccurate (see Figure 8).

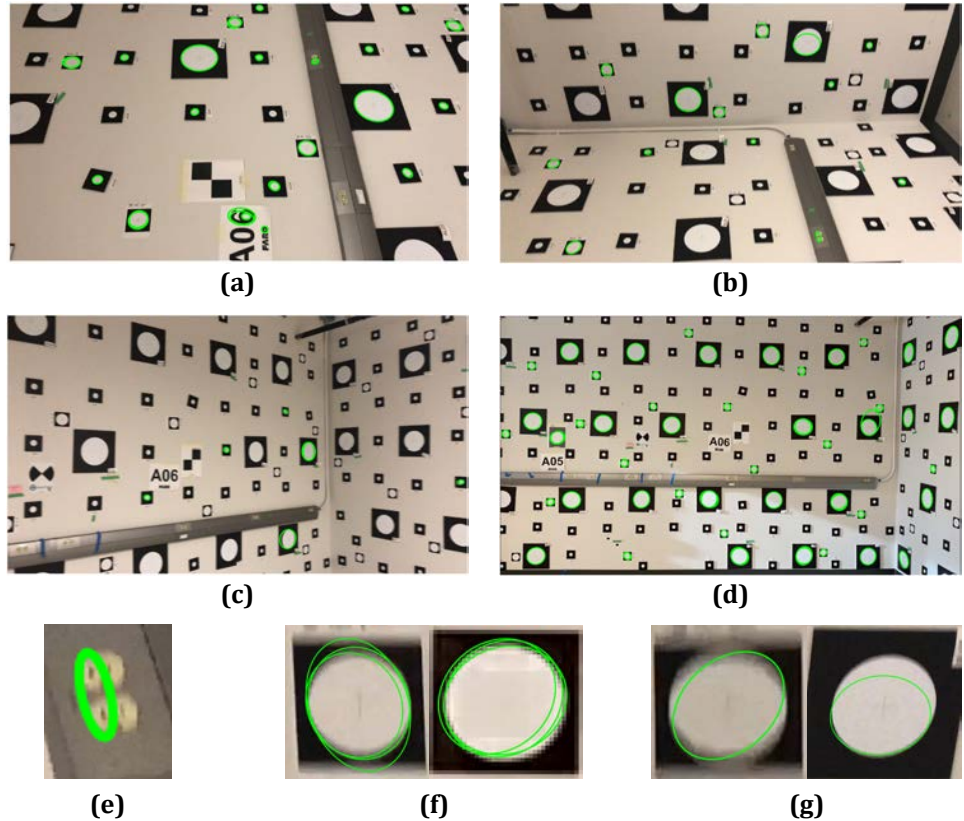

**Figure 17:** Detected ellipses (green) using the method of (Fornaciari et al., 2014); image of Figure: a) 16a; b) 16d; c) 16e; d) 16f. Examples of detection in accuracies: e) Type II detection error; f) multiple overlapping ellipses fitted to a single ellipse; and g) incorrect and biased ellipse fitting on connected points.

The results obtained by our method clearly outperformed the methods of Fornaciari, Patraucean, and Panagiotakis in both Type I errors (elliptic segments not detected as elliptical), and Type II errors (non-elliptic segments detected as elliptical). The overall ellipse detection precision, recall, accuracy, and F-measure (all thirty images combined) for all methods is reported in Table 3. Table 3 also includes the detection quality of our method without the robust ellipse detection (i.e. up until the results of Figure 3e). As illustrated, our method achieved an overall F-measure of approximately 2.3-, 1.5-, and 1.7-times that achieved using the methods of Fornaciari, Patraucean, and Panagiotakis, respectively. Our robust method, therefore, achieved around 50% better F-measure compared to the next best performing method, namely, Patraucean (Pătrăucean et al., 2017). It was also observed that using our robust ellipse fitting the detection precision was approximately 15% better than when robust fitting was not used (with no significant difference in recall).

It is important to note that our method without robust fitting still outperformed that of Fornaciari, Patraucean, and Panagiotakis. This is attributed mostly to the fact that chaining methods use local neighborhoods of edge points (particularly the curvature change) to decide on the elliptic or non-elliptic behavior of a specific pixel. Specifically, for the problem of ellipse fitting, if a small neighborhood is used to detect the elliptic behavior of a large ellipse, the pixels, especially closer to the minor axes, will locally behave close to linear. This is attributed to the anisotropic curvature change in ellipses, which is comparatively more rapid at the major axes than the minor axes (low-curvature bias problem (Zhang, 1997)). Hence, the thresholds used to classify elliptic and non-elliptic points is a function of the size of an ellipse as well as the relative location of a pixel on the ellipse, which are unknown a-priori. Subjective thresholds will almost guarantee that ellipses of particular sizes (such as the cases in Figures 17d and 18d, also observed and corroborated in Fig.13 of (Li, 2019)), and edge points of particular regions on an ellipse (resulting in poor ellipse fitting in Figures 17g and 18g) are disregarded. Our method, on the other hand, uses a top-down approach with the assumption that each connected segment either encompasses exactly one ellipse or does not contain an ellipse (due to the non-overlapping nature of the ellipses). Therefore, a decision is made on the elliptic behavior of the whole segment rather than a local pixels' neighborhood. In this manuscript, due to the introduction of a new accurate and robust elliptic point extraction and a systematically derived validation criteria and threshold, the non-overlapping ellipses were detected with high robustness to both Type I and Type II errors.

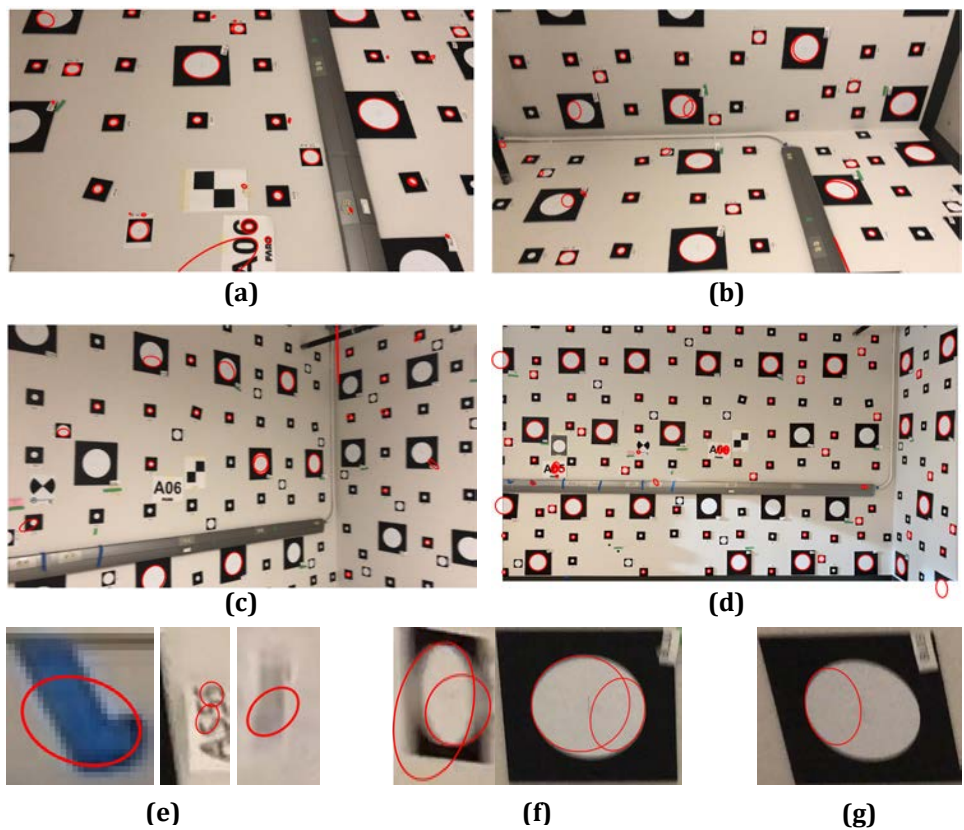

**Figure 18:** Detected ellipses (red) using the method of (Pătrăucean et al., 2017); image of Figure: a) 16a; b) 16d; c) 16e; d) 16f. Examples of detection in accuracies: e) Type II detection error; f) multiple overlapping ellipses fitted to a single ellipse; and g) incorrect and biased ellipse fitting on connected points.

**Table 3**: Overall precision, recall, accuracy, and F-measure (F1-score) for ellipse detection using each method

| Method | Precision | Recall | Accuracy | F-Measure |
|---|---|---|---|---|
| Ours | 98.5% | 100.0% | 99.5% | 99.3% |
| Ours- without robust fitting | 85.7% | 100.0% | 95.4% | 92.3% |
| Fornaciari (Fornaciari et al., 2014) | 29.7% | 73.8% | 70.6% | 42.4% |
| Patraucean (Pătrăucean et al., 2017) | 55.1% | 80.9% | 79.0% | 65.6% |
| Panagiotakis (Panagiotakis and Argyros, 2020) | 47.7% | 78.2% | 75.8% | 59.2% |

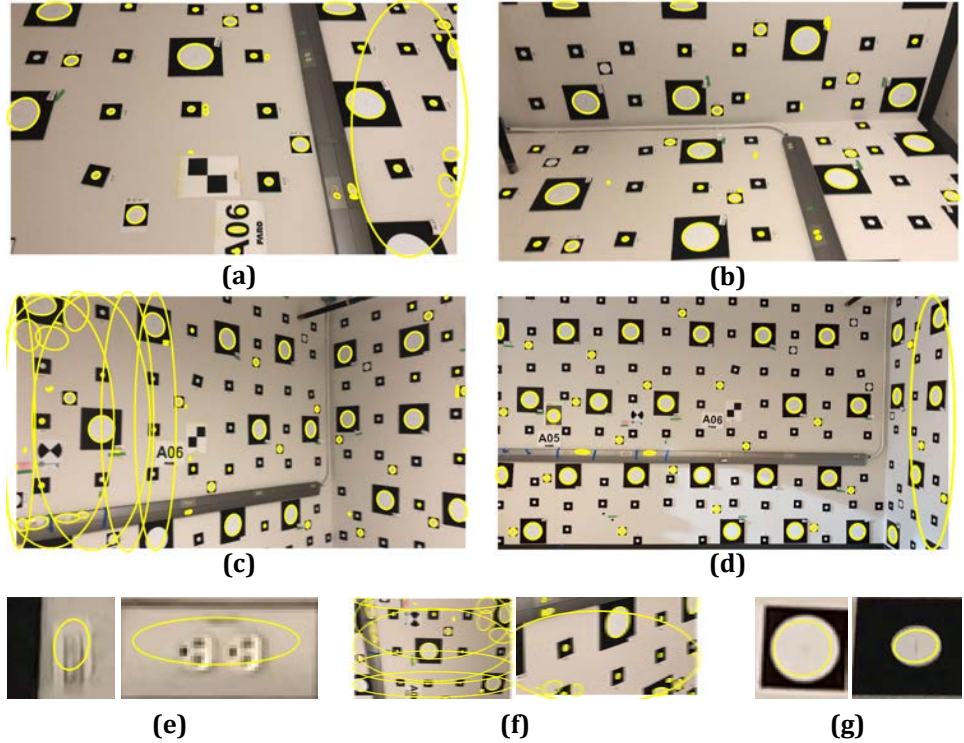

**Figure 19:** Detected ellipses (yellow) using the method of (Panagiotakis and Argyros, 2020); image of Figure: a) 16a; b) 16d; c) 16e; d) 16f. Examples of detection in accuracies: e) Type II detection error; f) multiple overlapping ellipses fitted to a single ellipse; and g) incorrect and biased ellipse fitting on connected points.

*5.4. Cylinder detection from point clouds*

*5.4.1. Cylinder detection under laboratory conditions*

    The feasibility of employing the proposed ellipse detection method to extract cylinders from point clouds was evaluated. Three manually selected planes, passing through six (in the cases of Figures 19a middle and right) to seven (in the cases of Figures 20a left) different mechanical pipes, were considered. Figure 20a shows the point cloud together with the three intersecting planes. Figure 20b shows the parallel projection of the points within 1mm of the planes (in the plane's coordinate system). The connected projected points are presented in Figure 20c. In Figure 20d, the extracted ellipses (from the connected segments) are shown in red (segments whose best fit ellipse meets the validation condition of equation 7), and green (segments whose best fit ellipse was detected by applying the robust ellipse detection of Algorithms 1-3). It can be observed that all ellipses, corresponding to the mechanical pipes, in the three plane arrangements were correctly detected. Using Algorithm 4, the cylinders were detected from the initial ellipse parameters. Figure 20e shows the final set of extracted cylinders for each plane, and its correspondence with the detected ellipse. Since all ellipses were detected correctly, the cylinders were also detected in all three plane arrangements.

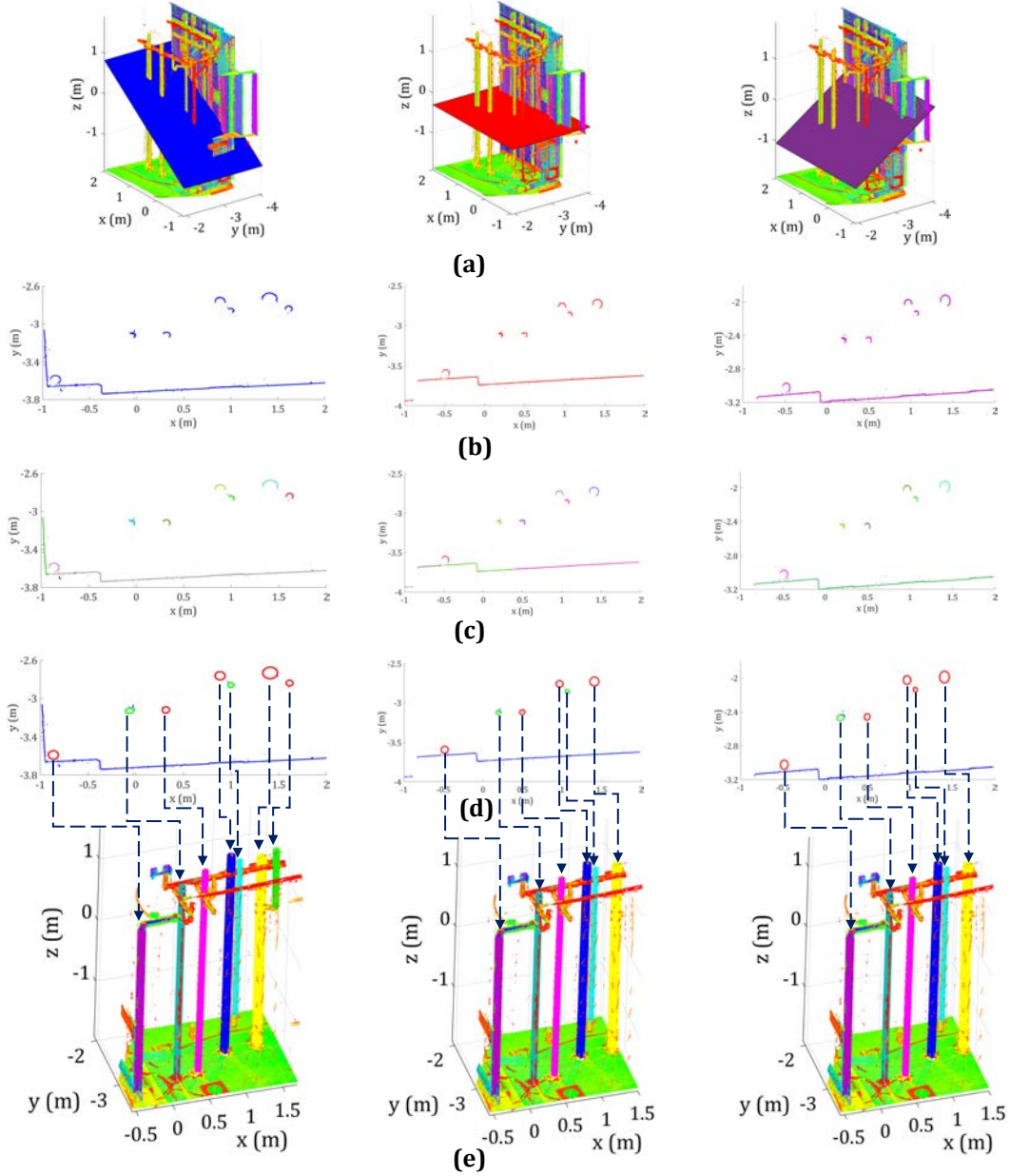

**Figure 20:** Detection of cylinders from point clouds of pipes captured under laboratory conditions: a) the three selected planes intersecting the mechanical pipes; b) projected points in the plane's coordinate system; c) connected components on the projected points; d) ellipse fitting (red) and robust ellipse fitting (green) on connected segments; and e) the extracted cylinders corresponding to the detected ellipses.

To provide a perspective on the accuracy of the initial cylinder parameter estimation using the ellipses, Table 4 shows the mean of the absolute error of the estimated center, radius, and cylinder axis between equation 14 and the ground truth. It can be observed that the ellipses detected from all plane arrangements estimated the ground cylinder's radius, center, and axis angle within 1.93mm, 2.14mm and 3.83° on average, respectively. In fact, based on empirical data presented in (Maalek et al., 2019), the robust cylinder extraction method effectively recovered the ground truth cylinder's axis even in cases with around 7° angular error between the initial and ground truth cylinders axis (see Fig. 3.3). Therefore, the initial cylinder's axis, estimated using the proposed method, was sufficient to detect all cylinders in the presented dataset.

**Table 4:** Mean of the cylinder parameter estimation errors between ground truth and ellipse fit for the cases presented in Figure 20

| Mean of Cylinder Parameter Errors | | | |
|---|---|---|---|
| **Cylinder Parameter Error** | **Figure 20-left** | **Figure 20-middle** | **Figure 20-right** |
| **Radius (mm)** | 1.93 | 1.00 | 0.31 |
| **Center (mm)** | 2.14 | 1.55 | 1.73 |
| **Axis Angle* (°)** | 3.37 | 2.36 | 3.83 |

\* Axis angle is the angle between the estimated axis and the ground truth axis

*5.4.2. Cylinder detection from industrial construction site*

This section, similar to the previous, presents the results of cylinder detection from point cloud, acquired from a portion of an industrial construction site. Three planes, passing through six, thirteen, and six cylinders for the cases presented in Figures 21a-left, 21a-middle, and 21a-right, respectively, were considered. Figures 21b, 21c, and 21d, shows the projected points, connected projected points, and detected non-overlapping ellipses, respectively, for the three planes. It can be observed that exactly six, thirteen and six ellipses were detected using our proposed non-overlapping ellipse detection, corresponding to the six, thirteen, and six cylinders that intersect the cylinders of Figures 21a-left, 21a-middle, and 21a-right, respectively. Following the findings of the laboratory experiment, the proposed method was found to correctly detected cylinders, even under adverse construction site conditions (Figure 213) in the presence of natural occlusions, such as cuboid-like columns and beams (projected as rectangles in Figure 21b). As a point of reference, Table 5 shows the mean cylinder parameter estimation errors between the initially estimated cylinder parameters (using equation 14), and the final set of cylinder parameters. Overall, the parameter estimation errors were slightly higher than the laboratory; however, they were still within the range, where the cylinder fitting of (Maalek et al., 2019) was shown to detect cylinders consistently.

**Table 5:** Mean of the cylinder parameter estimation errors between ground truth and ellipse fit for the cases presented in Figure 21

| Mean of Cylinder Parameter Errors | | | |
|---|---|---|---|
| **Cylinder Parameter Error** | **Figure 21-left** | **Figure 21-middle** | **Figure 21-right** |
| **Radius (mm)** | 0.44 | 3.18 | 2.71 |
| **Center (mm)** | 0.30 | 5.14 | 3.73 |
| **Axis Angle* (°)** | 6.77 | 2.61 | 2.83 |

\* Axis angle is the angle between the estimated axis and the ground truth axis

## 6. Discussion and Summary of Finding

Four categories of experiments were designed for: (i) comparative analysis of robust ellipse fitting; (ii) sensitivity analysis of ellipse validation criterion; (iii) comparison of non-overlapping ellipse detection from images; and (iv) cylinder extraction from point clouds. It was shown that random selection of 10,000 subsets of six points from simulated ellipses can produce parameter estimation errors of up to 300 times that of the worst subset obtained using our subsampling method (with roughly 62% of the random subsets achieving larger parameter estimation errors than ours). It was also observed that the proposed robust ellipse fitting outperformed four reliable methods, namely, the LMedS with random sampling, LMedS with our proposed sampling strategy, SARE (Thurnhofer-Hemsi et al., 2020) and Munoz (Muñoz-Pérez et al., 2014). It was also demonstrated that our robust method was considerably less impacted by an increase in noise levels as well as outlier ratio compared to LMedS, SARE and Munoz. For ten samples of real-world connected points of edge images and projected point clouds with an average outlier ratio of 58.6%, our method achieved an F-measure of roughly 1.6- and 1.7- times higher than LMedS using our sampling strategy and LMedS using random sampling, respectively.

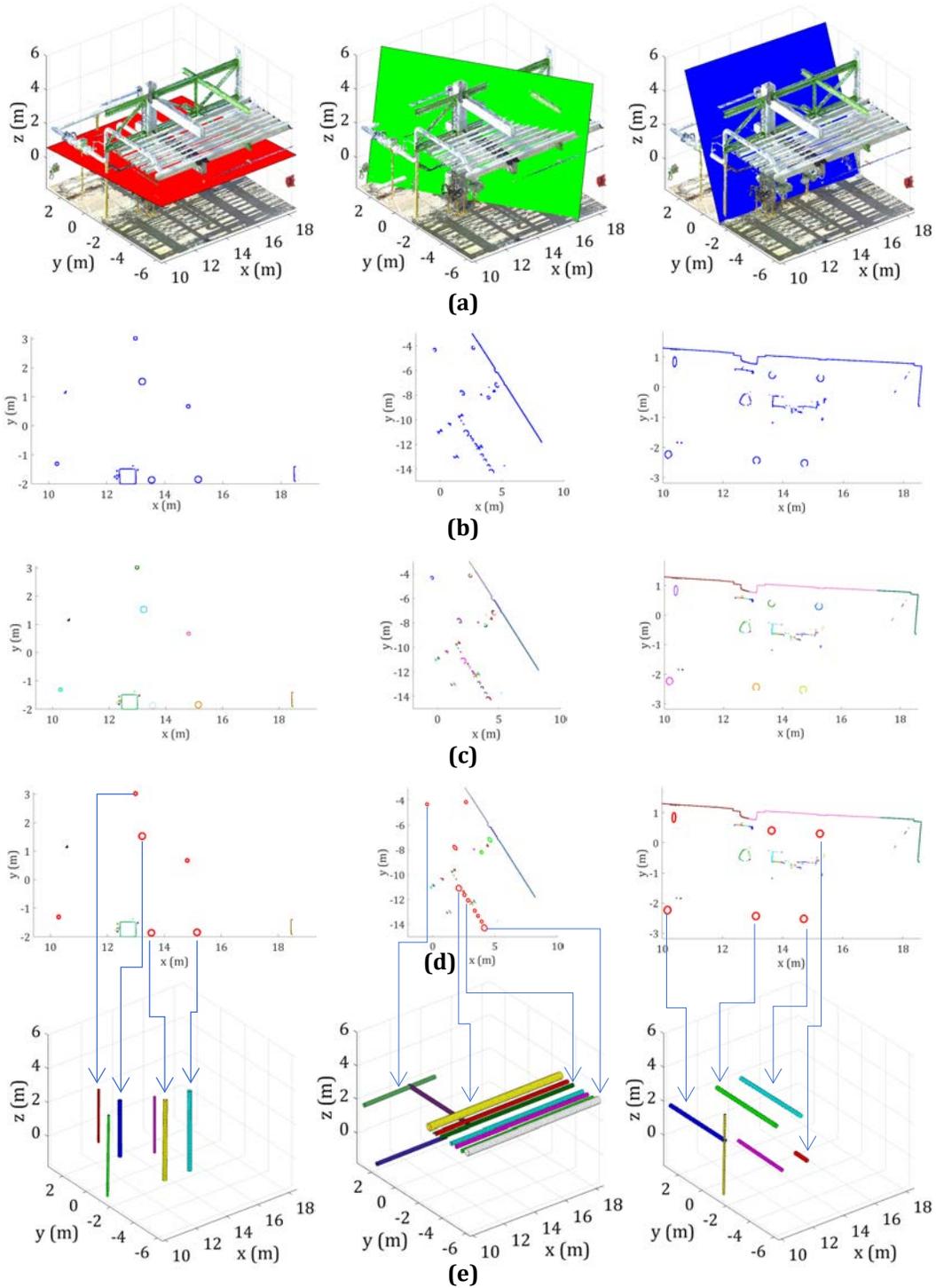

**Figure 21:** Detection of cylinders from point clouds of pipes captured under industrial construction site conditions: a) the three selected planes intersecting the mechanical pipes; b) projected points in the plane's coordinate system; c) connected components on the projected points; d) ellipse fitting (red) and robust ellipse fitting (green) on connected segments; and e) the extracted cylinders corresponding to the detected ellipses.

Sensitivity analysis of the ellipse validation criteria showed that the Euclidian ellipticity can differentiate ellipses from rectangles, more effectively than the moment invariants and elliptic variance metrics. Through many simulated ellipses and rectangles, the Euclidian ellipticity threshold to detect ellipses with low Type I and Type II errors was found to be between 0.955 and 0.965. This latter threshold range was validated and confirmed in thirty real-world images, containing multiple circular target fields. The thirty real-world images, taken from the calibration target field, were also used to evaluate the detection quality of the proposed method for non-overlapping ellipse detection, compared to the methods of Fornaciari, Patraucean, and Panagiotakis. It was shown that our method achieved an F-measure of 99.3%, which was around 2.3-, 1.5-, and 1.7- times that obtained using the methods of Fornaciari, Patraucean, and Panagiotakis, respectively.

Finally, the proposed procedure to extract cylinders from point clouds was evaluated on 2D points projected onto three random planes, intersecting multiple mechanical pipes, in two separate datasets. The two datasets provided the opportunity to assess the effectiveness of the proposed methods in both controlled laboratory and actual industrial construction site conditions. All ellipses formed through the projected points of cylindrical pipes were detected in all plane arrangements of both datasets. Overall, the best fit ellipse parameters predicted the ground truth cylinder radius, center, and axis angle within 3.18mm, 5.14mm, and 6.77°, respectively, on average. As a result, the initially estimated cylinder parameters were close enough in all cases to recover the final cylinder parameters.

## 7. Summary of Methods and Avenues for Future Exploration

This manuscript provided a collection of new algorithms for the accurate and robust extraction of non-overlapping ellipses from 2D points, applicable to both circular target extraction from images and cylinder extraction from point clouds. Predicated on the comprehensive survey of previous related work (provided in Sections 1 and 2), the successful algorithm requires a robust method for detecting ellipses in the presence of outliers, and a systematic ellipse validation criterion to detect ellipses from non-elliptic segments. These categories of requirements were addressed through: (i) a new Monte Carlo-based method for robust identification of elliptic points in the presence of outliers, which consists of new strategies for subsample selection, robust ellipse extraction, and ellipse validation; and (ii) a new connected segment validation criterion, which uses the Euclidian ellipticity measure with a systematically derived threshold. Other than the immediate application of the proposed method for extraction of circular (spherical) objects from images, a procedure was developed to extract cylinders from point clouds, using the extracted ellipses of the points projected onto an intersecting plane. The effectiveness of these methods was thoroughly examined in both simulated edge points of real-world images, taken from a calibration room, as well as 3D point clouds, acquired from professionally installed mechanical mock pipes as well as an industrial construction site.

The results obtained through the various experimental evaluations presented in this manuscript suggest potential for the generic applicability of the proposed methods for detection of non-overlapping ellipses from 2D edge points to address the problem of circular target extraction from images and cylinder detection from point clouds. The following points may also provide avenues for future exploration and expansion:

1- The robust ellipse fitting was proposed to help neutralize the incorrect image binarization. Currently, ongoing initiatives, such as the document image binarization competition (DIBC) (Pratikakis et al., 2017), focus on improving the accuracy of image binarization. Therefore, an important topic for future research is to quantify the impact of different image binarization methods on the robustness of the proposed ellipse detection method. In line with such expansions, the impact of Gaussian noise and blurring can be investigated on both the robust ellipse fitting along with the ellipse validation criteria.

2- The proposed methods solely focused on non-overlapping ellipse detection from 2D points. Their extension to overlapping ellipses along with a combined approach with advanced computer vision methods such as supervised learning, might also be an interesting topic for investigation.
3- The proposed cylinder detection method requires the definition of a plane, which passes through the cylinders. Future research can focus on performing pre-processing to determine the minimum number of planes along with their attributes to ensure adequate point density and coverage of all cylinders in the point cloud.
4- In line with cylinder detection from point clouds, comparison of the performance of the proposed method with methods proposed in (Birdal et al., 2020), (Nurunnabi et al., 2019), and (Bergamasco et al., 2020) would be worth exploring.

**Acknowledgement**

The authors wish to acknowledge the support provided by the MJS Mechanical Ltd. and Michael Baytalan for their cooperation and professional installation of the mechanical mock pipes for the purpose of the experiments, presented in this manuscript. This research project was partly funded by the Natural Sciences and Engineering Research Council (NSERC) of Canada (542980 - 19) and Alberta Innovates (G2020000051).